\documentclass[letterpaper, 10 pt, conference]{ieeeconf}  

\IEEEoverridecommandlockouts                              
\usepackage[colorlinks]{hyperref}

\usepackage{amsmath} 
\usepackage{amssymb}  
\usepackage{listings}
\usepackage{subcaption}
\usepackage{graphicx}
\usepackage{tikz}

\usetikzlibrary{positioning}

\usepackage{pifont}
\newcommand{\cmark}{\ding{51}}%
\newcommand{\xmark}{\ding{55}}%

\usepackage{multirow}
\usepackage{rotating}
\usepackage{booktabs}

\usepackage{cases}
\usepackage{mathtools}

\usepackage{mysymbols}
\usepackage{epigraph}

\newcommand{\nMesh}{18} 
\newcommand{\nMeshWoFRLApt}{12} 
\newcommand{\nRoom}{35}
\newcommand{\nFRLApt}{6}
\newcommand{\nOffice}{5}
\newcommand{\nAptRooms}{3}
\newcommand{\nAptMultiRoom}{2}
\newcommand{\nClass}{88} 

\begin{document}

\title{The Replica Dataset: A Digital Replica of Indoor Spaces \\[5pt] 
{\large \href{url}{https://github.com/facebookresearch/Replica-Dataset}}\\[-5pt]}

\author{%
Julian Straub$^1$,  
Thomas Whelan$^1$, 
Lingni Ma$^1$,
Yufan Chen$^1$,
Erik Wijmans$^{2,3}$,
Simon Green$^1$, 
Jakob J. Engel$^1$,  \\
Raul Mur-Artal$^1$, 
Carl Ren$^1$,
Shobhit Verma$^1$,
Anton Clarkson$^1$,
Mingfei Yan$^1$,
Brian Budge$^1$, 
Yajie Yan$^1$, \\
Xiaqing Pan$^1$, 
June Yon$^1$,
Yuyang Zou$^1$,
Kimberly Leon$^1$,
Nigel Carter$^1$,
Jesus Briales$^1$,  
Tyler Gillingham$^1$,  \\
Elias Mueggler$^1$,
Luis Pesqueira$^1$,
Manolis Savva$^{2,4}$,
Dhruv Batra$^{2,3}$,
Hauke M. Strasdat$^1$,  \\
Renzo De Nardi$^1$, 
Michael Goesele$^1$,
Steven Lovegrove$^1$, 
Richard Newcombe$^1$ \\[10pt]
$^1$Facebook Reality Labs, $^2$Facebook AI Research, $^3$Georgia Institute of Technology, $^4$Simon Fraser University
}

\twocolumn[{%
\renewcommand\twocolumn[1][]{#1}%
{\maketitle}%
{\begin{center}
\includegraphics[width=0.46\linewidth]{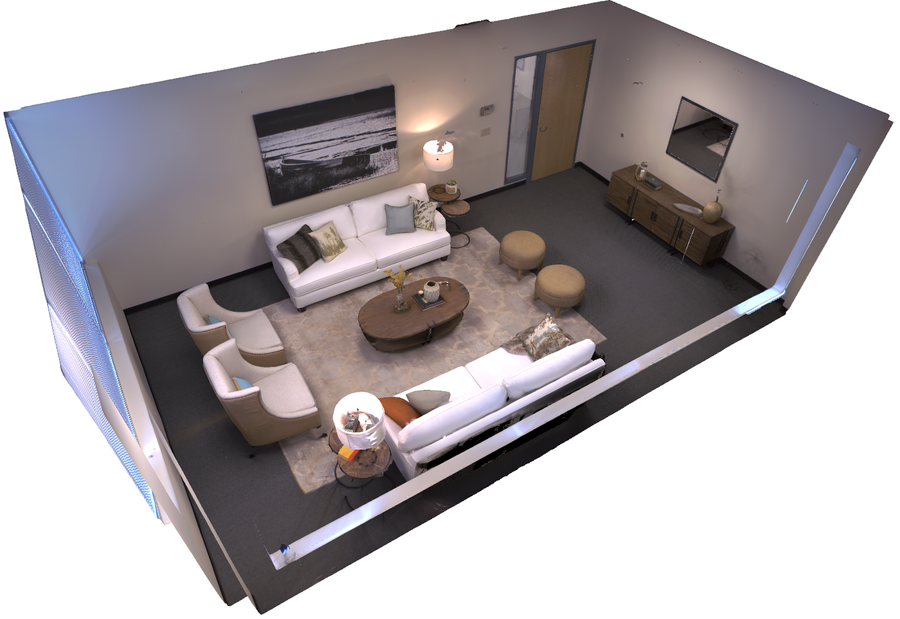}\hspace{0.5cm}
\includegraphics[width=0.46\linewidth]{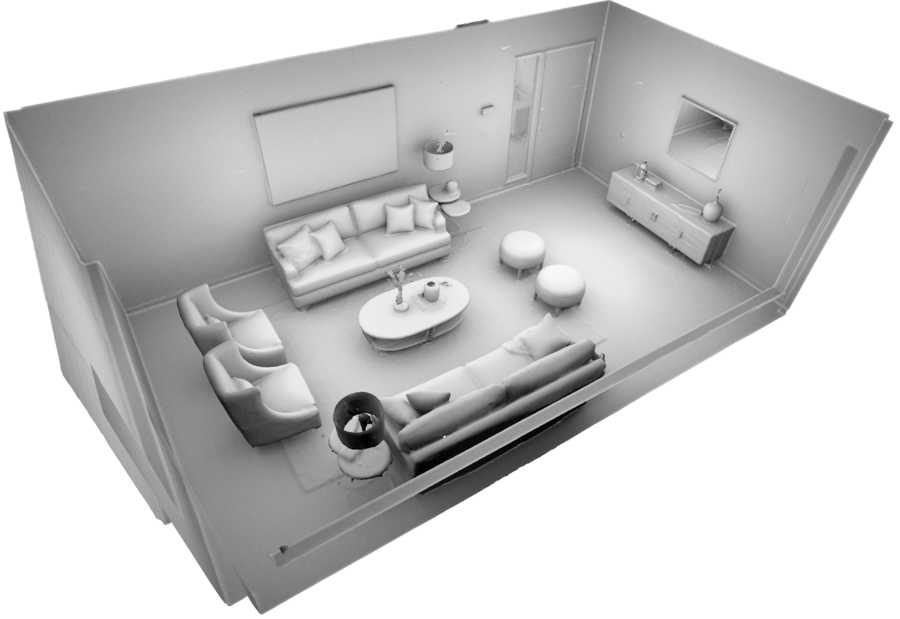}\\
\includegraphics[width=0.46\linewidth]{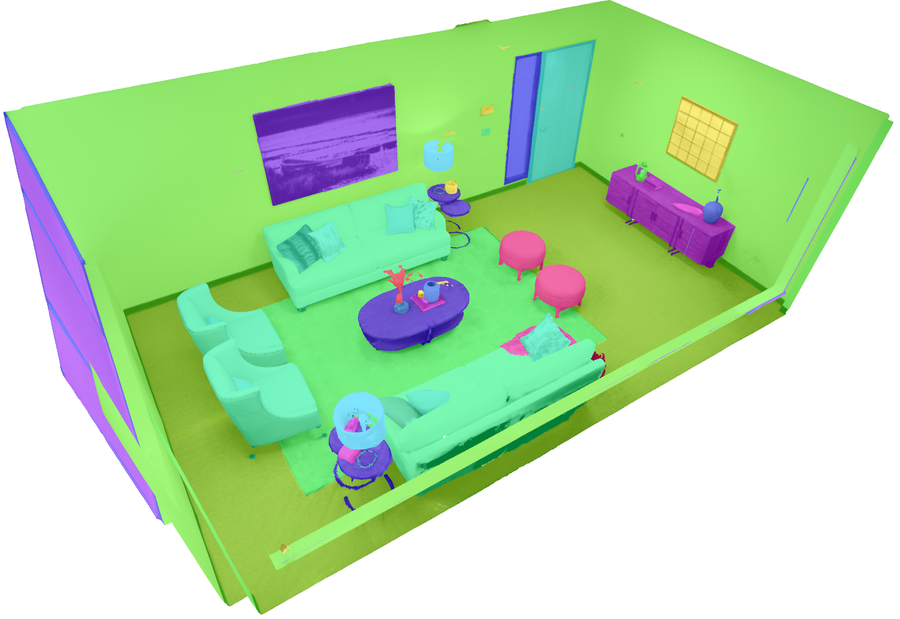}\hspace{0.5cm}
\includegraphics[width=0.46\linewidth]{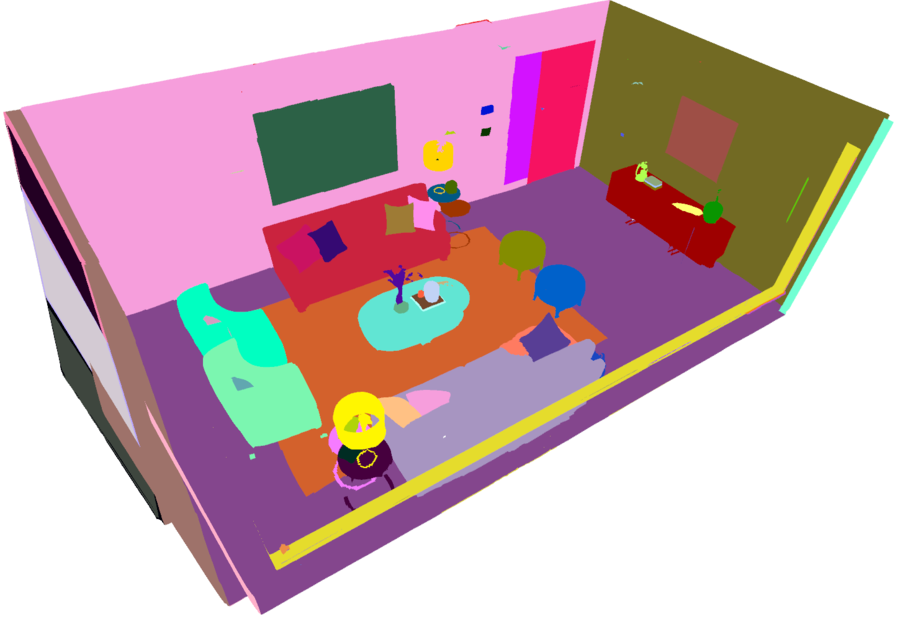}\\[5pt] 
  \captionof{figure}{The Replica dataset consists of \nMesh{} high resolution
  and high dynamic range (HDR) textured reconstructions with semantic class and
  instance segmentation as well as planar mirror and glass reflectors.} \label{fig:teaser}
\end{center}}%
}]

\thispagestyle{empty}
\pagestyle{empty}

\begin{abstract}
  We introduce Replica, a dataset of \nMesh{} highly photo-realistic 3D indoor
  scene reconstructions at room and building scale. Each scene consists of a
  dense mesh, high-resolution high-dynamic-range (HDR) textures, per-primitive
  semantic class and instance information, and planar mirror and glass reflectors.
The goal of Replica is to enable machine learning (ML) research that relies on visually, geometrically, and semantically realistic generative models of the world -- for instance, egocentric computer vision, semantic segmentation in 2D and 3D, geometric inference, and the development of embodied agents (virtual robots) performing navigation, instruction following, and question answering.
Due to the high level of realism of the renderings from Replica, there is hope that ML systems trained on Replica may transfer directly to real world image and video data.
Together with the data, we are releasing a minimal C++ SDK as a starting point
  for working with the Replica dataset. In addition, Replica is
  `Habitat-compatible', \ie can be natively used with AI
  Habitat~\cite{habitat19arxiv} for training and testing embodied agents. 
\end{abstract}

\section{Introduction}

\setlength{\epigraphwidth}{0.95\columnwidth}
\renewcommand{\textflush}{flushepinormal}
\renewcommand{\epigraphflush}{flushleft}
\epigraph{
\emph{If the organism carries a ``small scale model'' of external reality and of its own possible actions within its head, 
it is able to try out various alternatives, conclude which is the best of them, react to future situations before they 
arise, utilize the knowledge of past events in dealing with the present and future, and in every way to react in a 
much fuller, safer, and more competent manner to the emergencies that face it.}}
{Kenneth Craik~\cite{craik43} via Sutton and Barto~\cite{sutton81}}

Replicating real physical spaces in their full fidelity in a digital form is a longstanding goal across multiple areas in science and engineering. 
Digitizing real environments has many future use cases, such as virtual telepresence. 
The combination of replicas of real environments with powerful 
simulators such as AI Habitat~\cite{habitat19arxiv} enables scalable machine learning
that may yield models that can be directly deployed in the real world to perform tasks like embodied navigation~\cite{Anderson2018-Evaluation}, instruction following~\cite{Anderson2018-Language}, and question answering~\cite{embodiedqa}.
Via parallelization, reality simulators enable faster-than-realtime and more
scalable training of AI agents in comparison with training real robots in the
wild. 
Additionally, simulation from Replica can be leveraged in egocentric computer
vision, semantic segmentation in 2D and 3D and geometry inference.
More realistic replicas lead to more realistic virtual
telepresence, more accurate computation over them, and a smaller 
domain gap between simulation and reality.

\begin{figure}
    \centering
    \includegraphics[width=\linewidth]{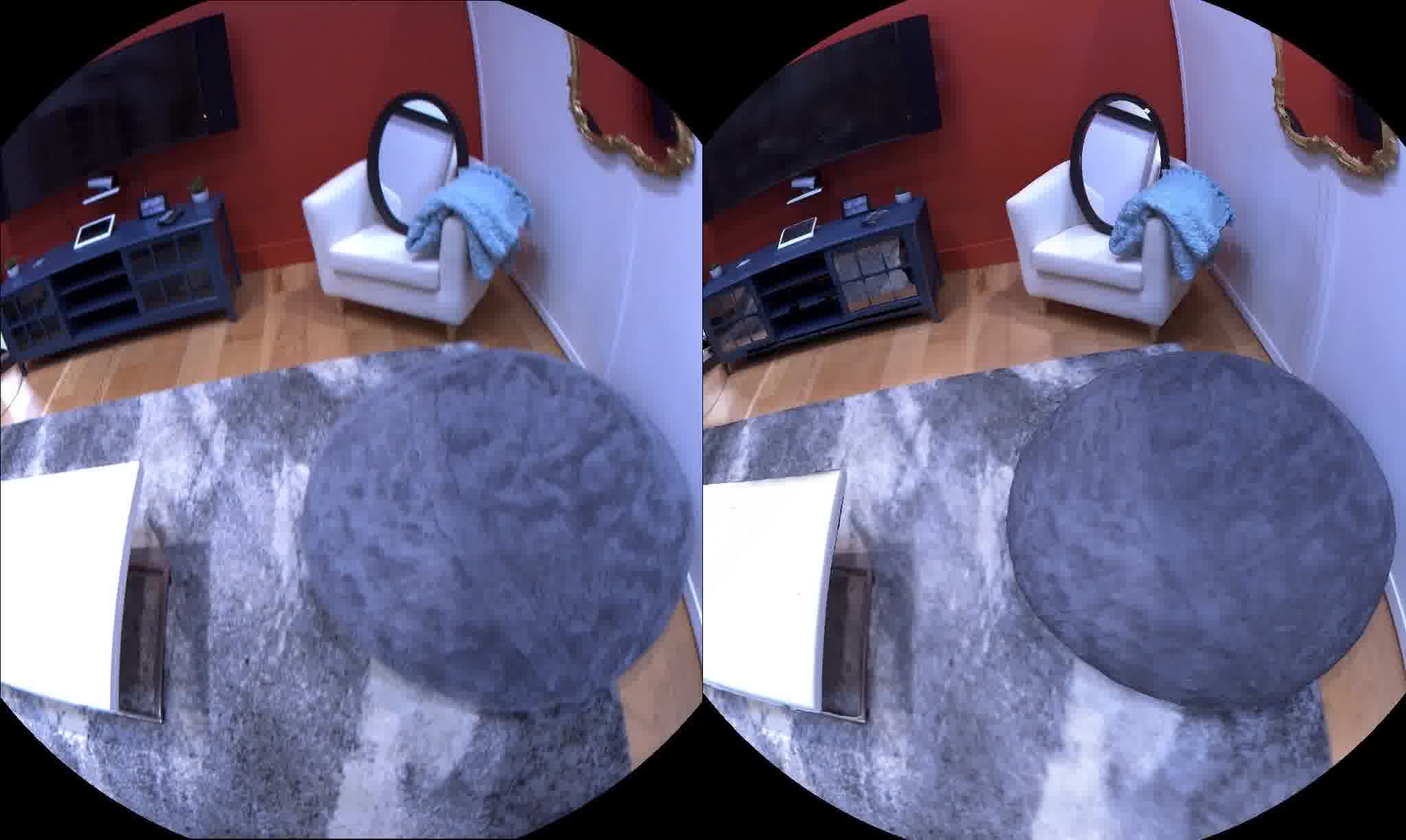} \\[5pt]

    \includegraphics[width=\linewidth]{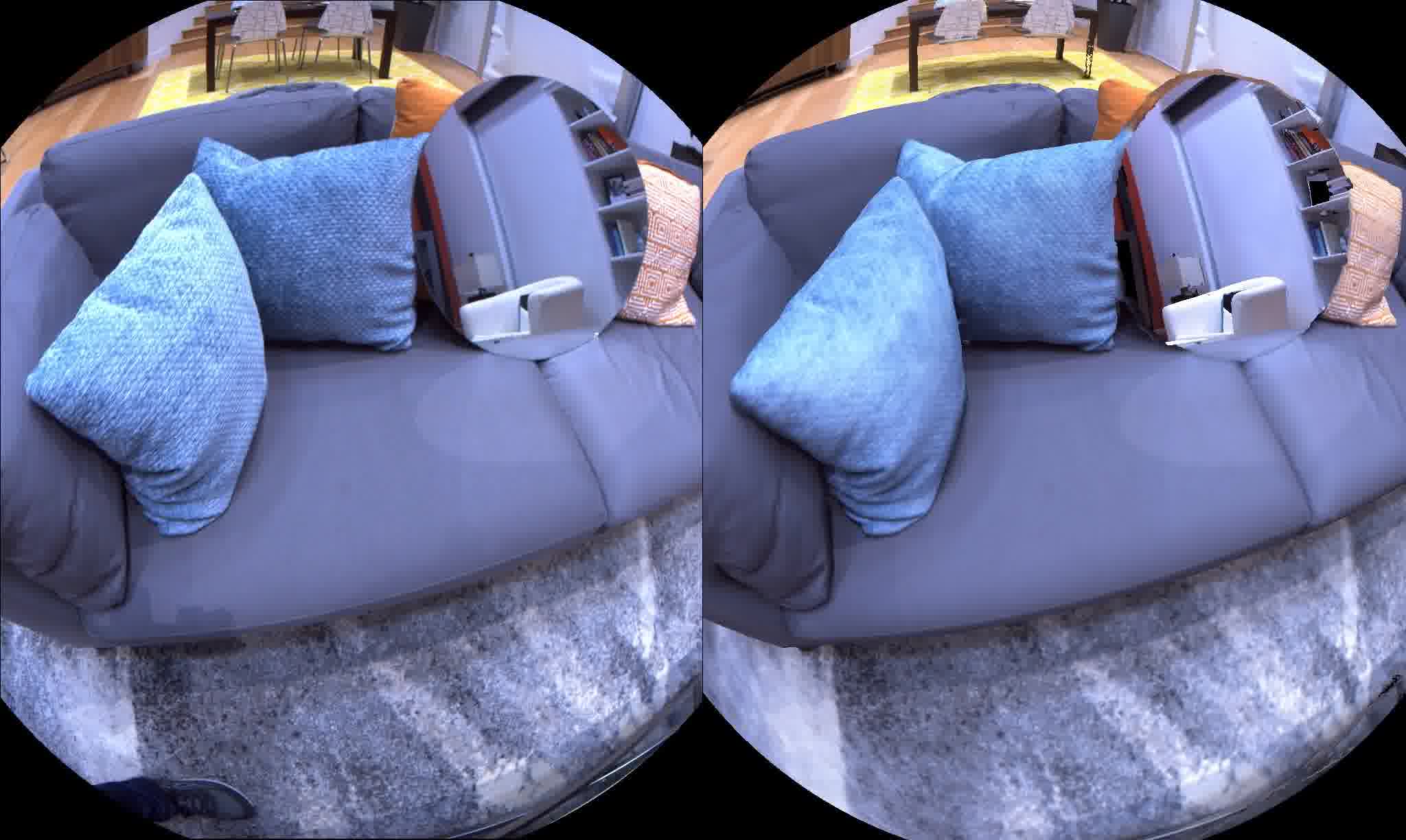} \\[5pt]

    \includegraphics[width=\linewidth]{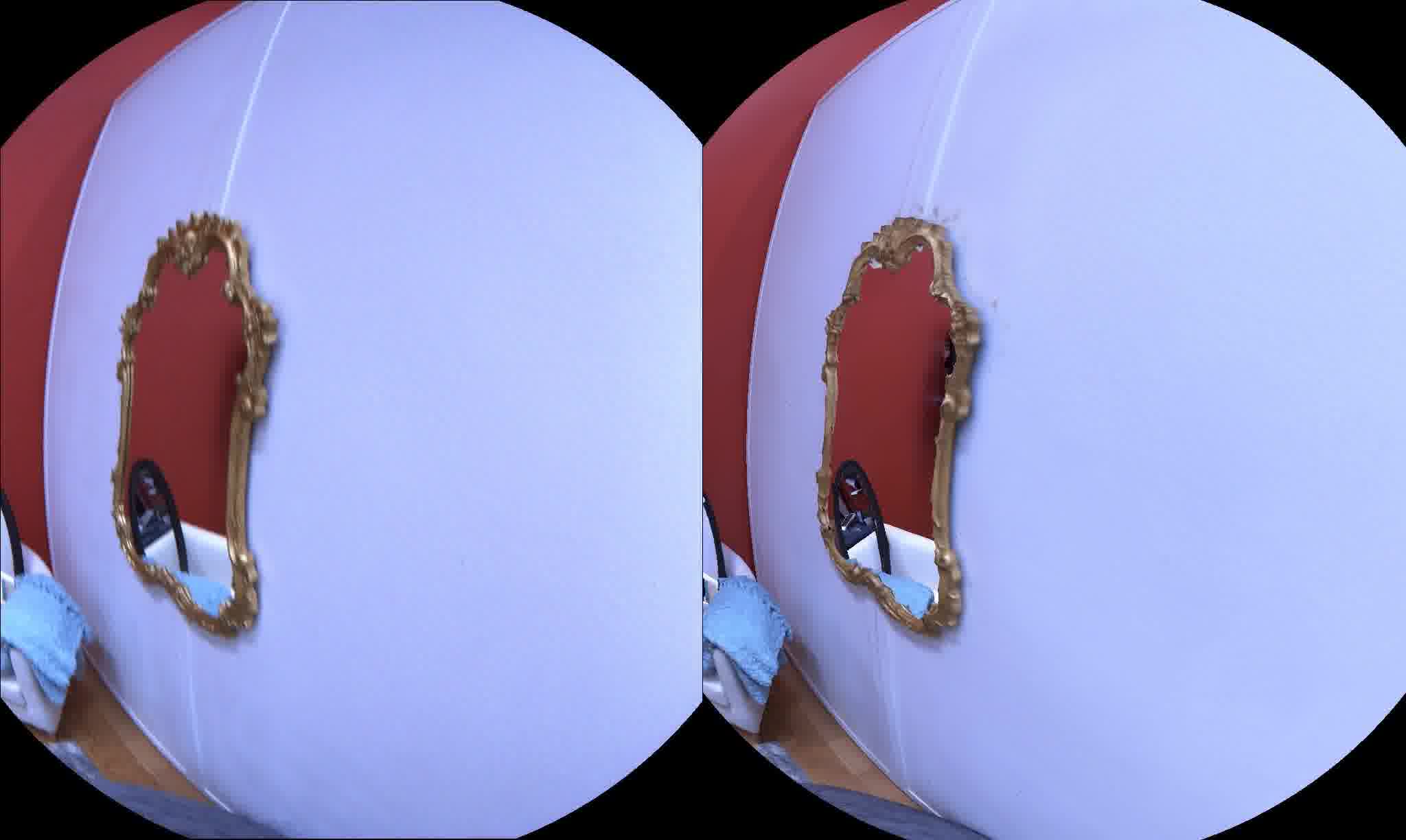} \\[5pt]

    \includegraphics[width=\linewidth]{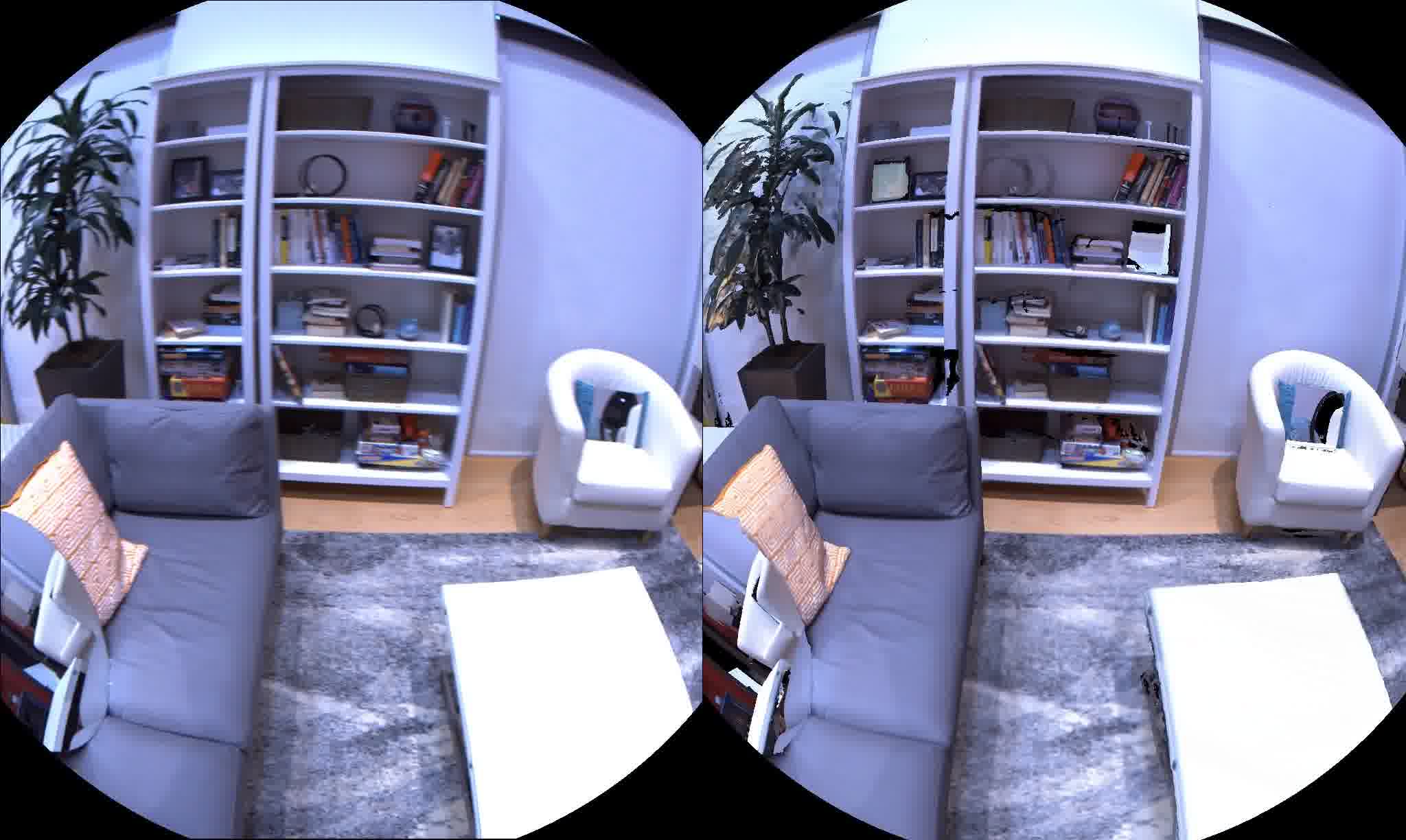}

    \caption{Replica `Turing Test': One column shows the raw RGB images captured in these spaces, the other column shows renderings from Replica (from the
    same camera pose). Can you tell which column shows `real' images and which column shows renderings? Find the answer in Sec.~\ref{sec:dataset}.}
    \label{fig:photometric}
\end{figure}

Datasets such as ImageNet~\cite{krizhevsky_nips12}, COCO~\cite{mscoco}, and VQA~\cite{antol_iccv15} have helped advance research in computer vision and multimodal AI problems.
With the Replica dataset we aim to unlock research into AI agents and assistants that can be
trained in simulation and deployed in the real world. The key distinction of Replica \wrt  
these image-based static datasets is that Replica scenes allow for \emph{active} perception since the
3D assets allow generating views from anywhere inside the model. This enables
the next generation of embodied AI tasks such as those studied in the AI Habitat platform~\cite{habitat19arxiv}. 
Compared to other 3D datasets such as Matterport 3D~\cite{Chang2017} and
ScanNet~\cite{dai2017scannet},
Replica achieves \emph{significantly} higher levels of realism -- we encourage you to take the Replica Turing Test in Fig.~\ref{fig:photometric}. 
Moreover, Replica introduces high dynamic range
(HDR) textures as well as renderable planar mirror and glass reflectors as can
be seen in the comparison of raw RGB capture with renders from the model in
Fig.~\ref{fig:photometric}. 
The Replica dataset contains \nMesh{} scenes of various real world environments. As
shown in Fig.~\ref{fig:teaser}, we provide a dense mesh, high resolution and HDR textures, semantic class and instance annotation of each primitive, and glass and mirror reflectors. 
The Replica dataset includes a variety of scene types as well as a large range of object instances from \nClass{} semantic classes to facilitate interesting machine learning tasks. It also contains \nFRLApt{} scans of the same indoor space with different furniture configurations that show different snapshots in time of the same space.

\begin{table*}[]
\centering
\setlength\tabcolsep{5pt}
\renewcommand{\arraystretch}{1.5}
\resizebox{\textwidth}{!}{
\begin{tabular}{@{}c p{1.5in} c c c c c @{}}
\toprule 
& & Replica & Matterport 3D (MP3D)  & ScanNet & Stanford 2D-3D-S & Gibson \\
\cmidrule(rl){3-3} \cmidrule(rl){4-4} \cmidrule(rl){5-5} \cmidrule(rl){6-6} \cmidrule(rl){7-7} 
\multirow{2}{*}{\rotatebox[origin=c]{90}{Scale} $\begin{dcases} \\ \\ \end{dcases}$} 
  & \# scenes & \nMesh{} & 90 & 1513 & 6    & 572 \\
  & \# rooms & \nRoom{} & 2056 & 707 & 270  & ? \\
\cmidrule{2-7}
\multirow{2}{*}{\rotatebox[origin=c]{90}{Res.} $\begin{dcases} \\ \\ \end{dcases}$}
  & color res.  $\left[ \tfrac{\text{pixel}}{m^2} \right]$ & 92k & 97k & 20k & $\approx$ MP3D & $\approx$ MP3D \\
  & geometry res. $\left[\tfrac{\text{primitives}}{m^2} \right]$ & $6$k &  $0.7$k & $20$k & $\approx$ MP3D & $\approx$ MP3D \\ 
\cmidrule{2-7}
\multirow{2}{*}{\rotatebox[origin=c]{90}{Fidelity} $\begin{dcases} \\ \\ \end{dcases}$}
& HDR textures  & \cmark & \xmark & \xmark  & \xmark & \xmark \\
& reflectors      & \cmark & \xmark & \xmark & \xmark & \xmark \\
\cmidrule{2-7}
\multirow{2}{*}{\rotatebox[origin=c]{90}{Labels} $\begin{dcases} \\ \\ \end{dcases}$}
  & semantic classes & \nClass{} & 40  & $\approx 1000$  & 13 & - \\
& semantic annotation & 3D Paint & 3D Felsenszwalb  & 3D Felsenszwalb & 3D & - \\
\bottomrule
\end{tabular}
}
  \caption{Comparison of reconstruction-based 3D scene datasets. 
  We estimate color and geometry resolution for each dataset as the number of
  pixels and mesh primitives respectively per $m^2$.
  Note that for
  all metrics we used the meshes that were semantically annotated and report
  median values. 
  \label{tab:related}}
\end{table*}



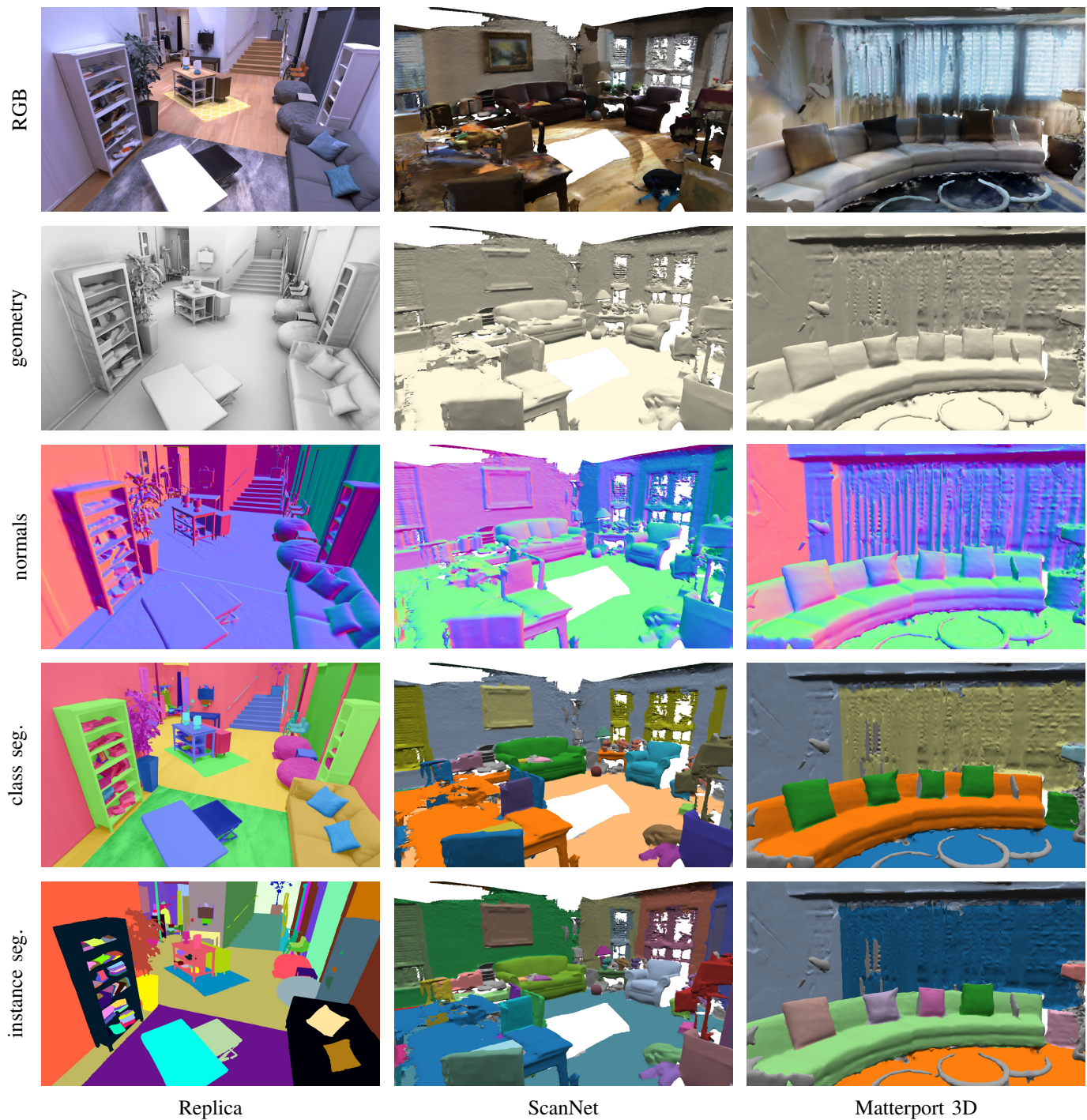
\begin{figure*}
    \centering
\begin{tikzpicture}
  \node (11) {
    \scalebox{1}[-1]{\includegraphics[width=0.32\linewidth,trim=130 0 0 0,clip]{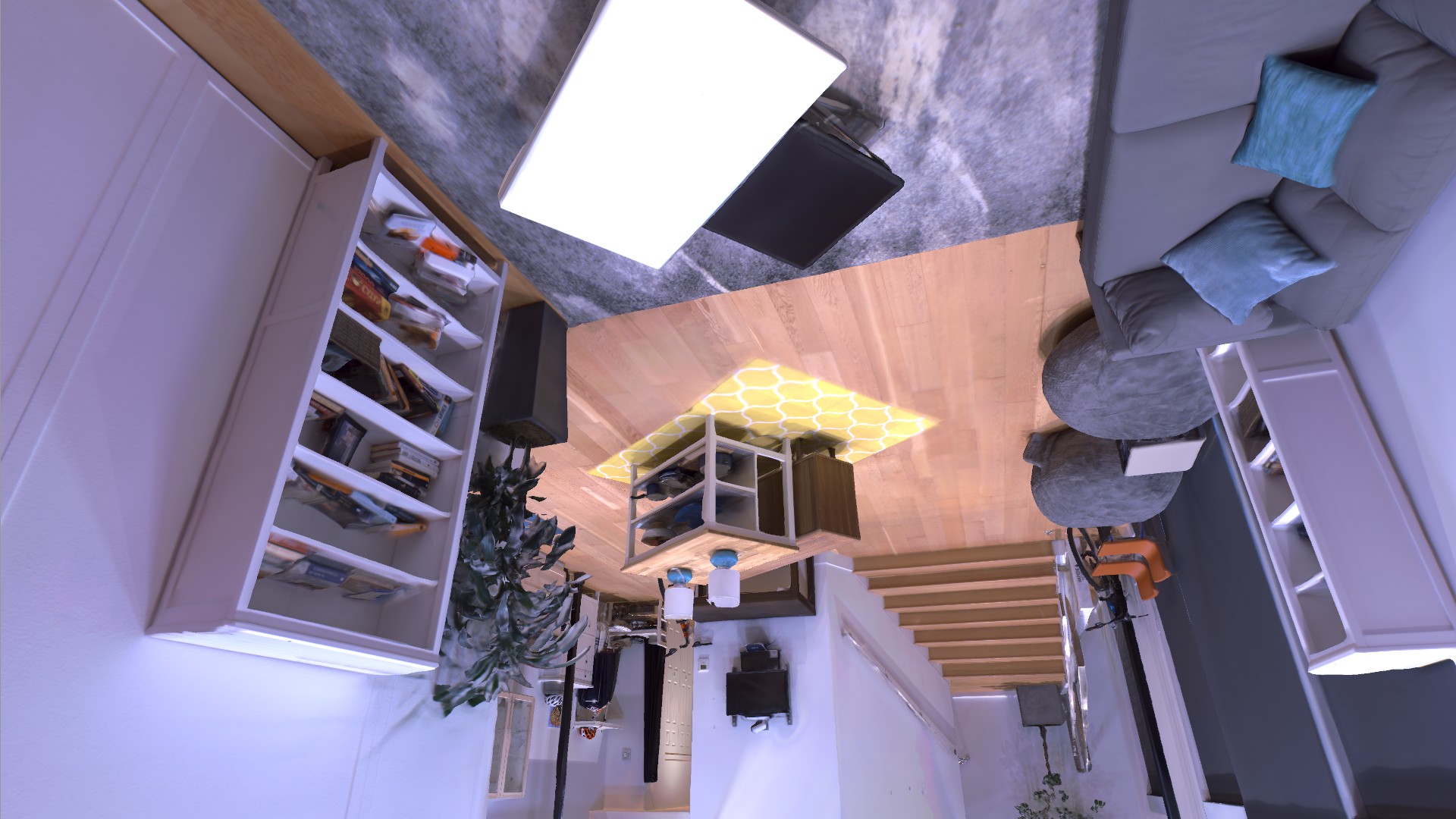}}
  };
  \node[right = 0ex of 11] (12) {
    \includegraphics[width=0.32\linewidth,trim=0 0 0 17,clip]{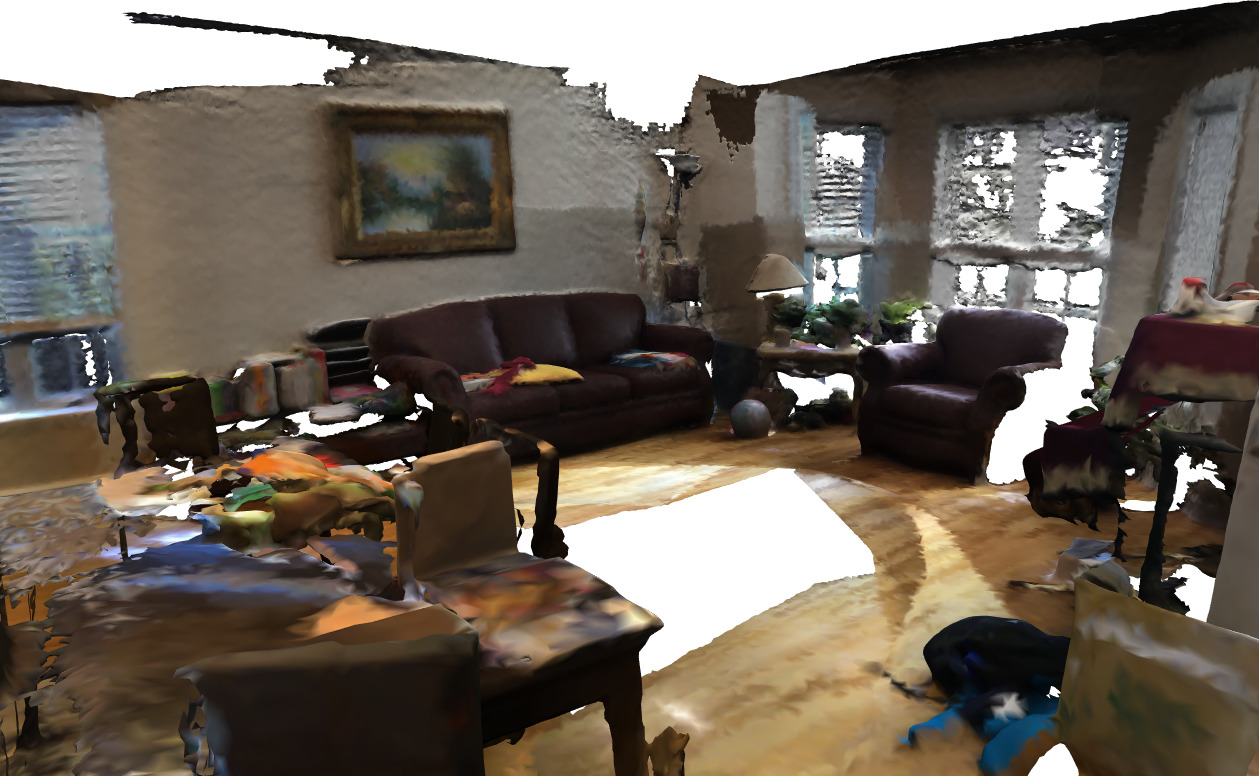}
  };
  \node[right = 0ex of 12] (13) {
    \includegraphics[width=0.32\linewidth,trim=0 28 0 93,clip]{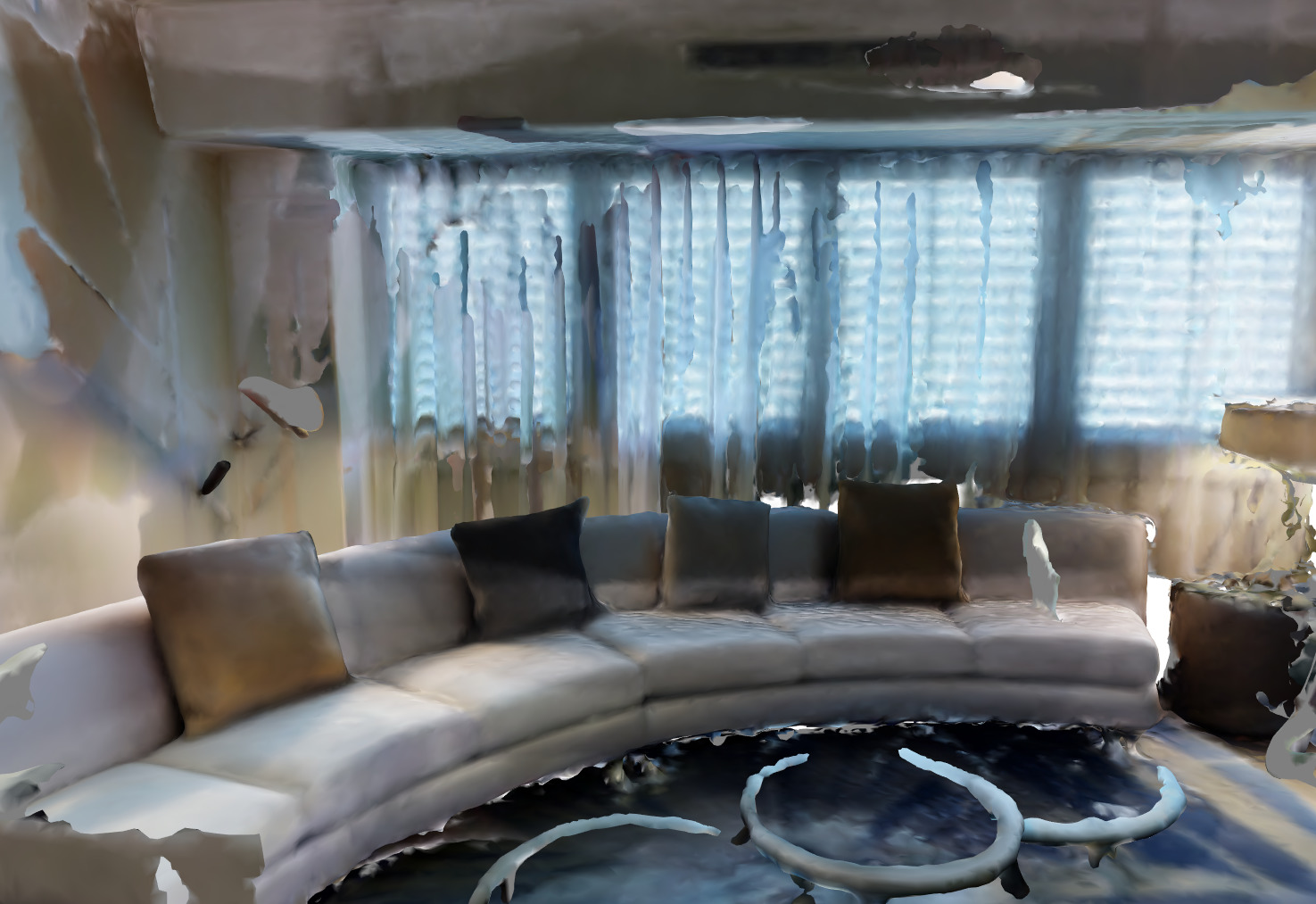}
  };

  \node[below = 0ex of 11] (21) {
    \scalebox{1}[-1]{\includegraphics[width=0.32\linewidth,trim=130 0 0 0,clip]{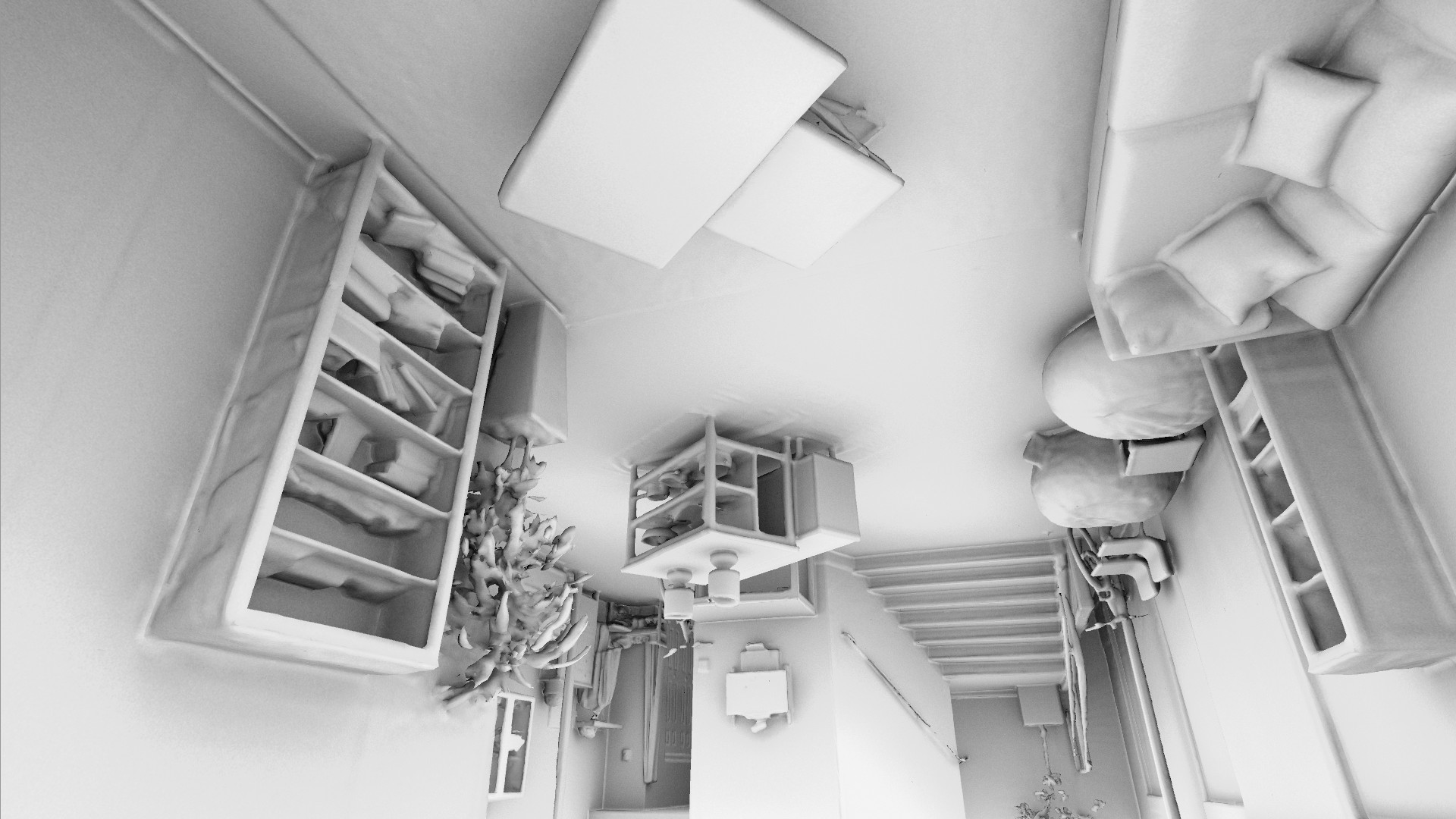}}
  };
  \node[right = 0ex of 21] (22) {
   \includegraphics[width=0.32\linewidth,trim=0 0 0 17,clip]{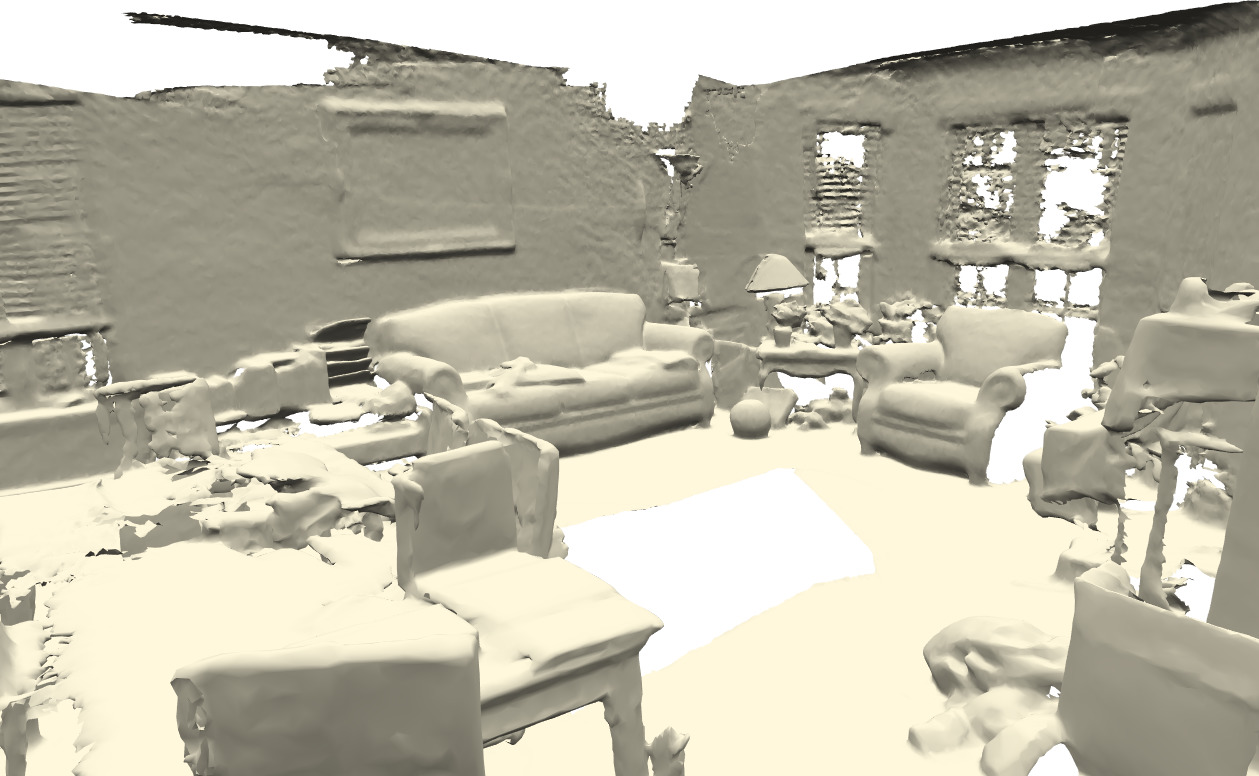}
  };
  \node[right = 0ex of 22] (23) {
    \includegraphics[width=0.32\linewidth,trim=0 28 0 93,clip]{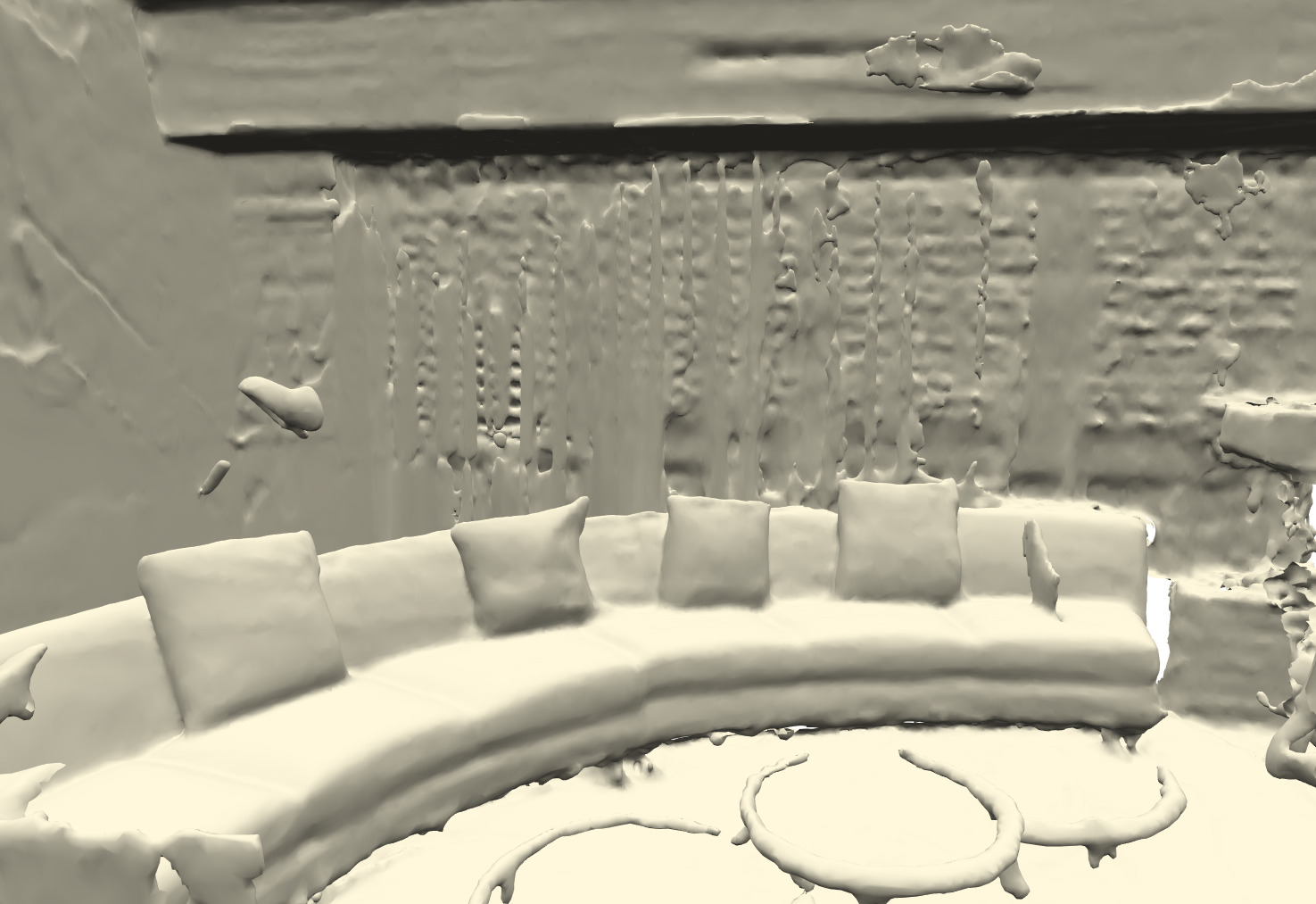}
  };
    
  \node[below = 0ex of 21] (31) {
    \scalebox{1}[-1]{\includegraphics[width=0.32\linewidth,trim=130 0 0 0,clip]{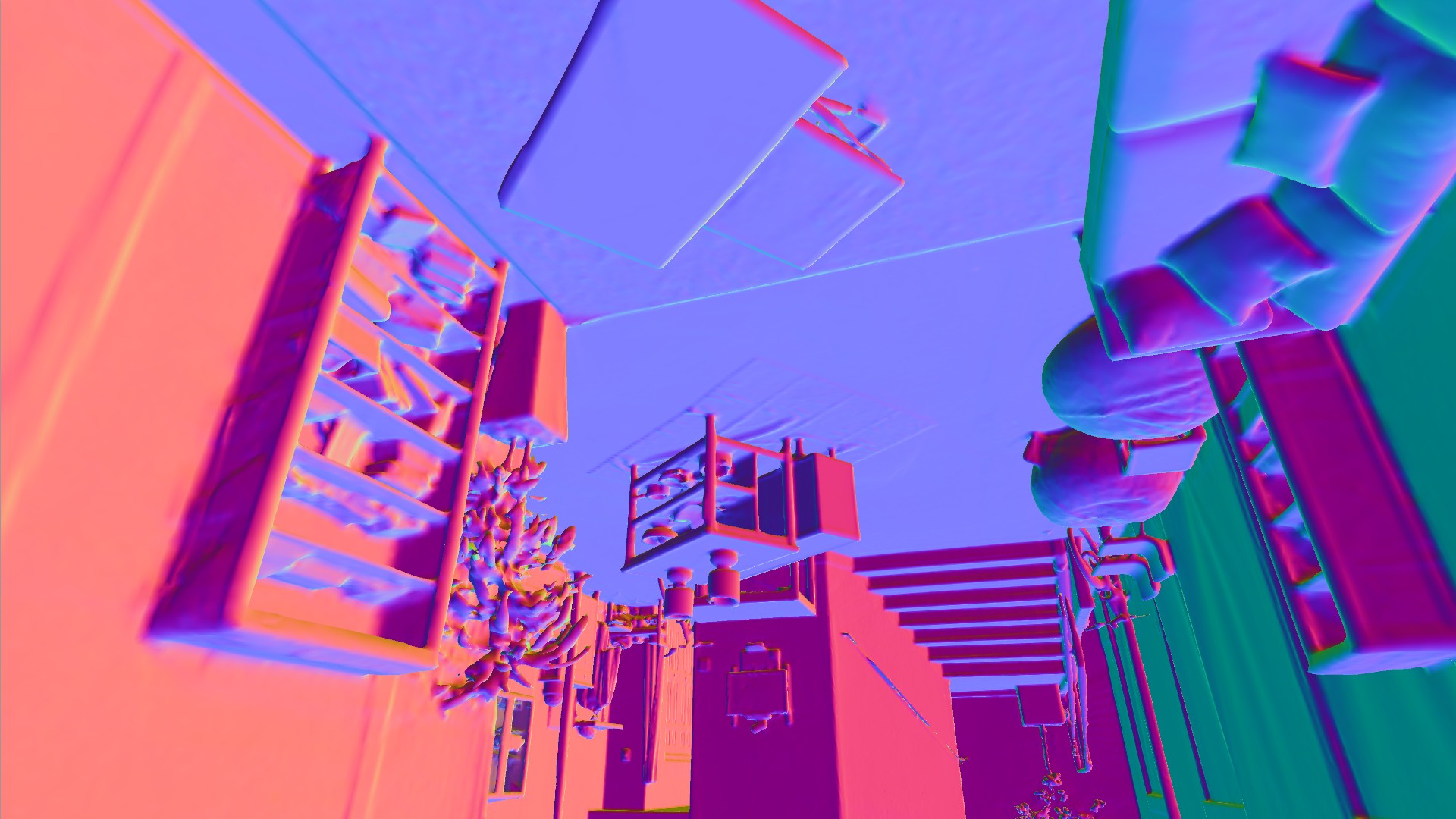}}
  };
  \node[right = 0ex of 31] (32) {
    \includegraphics[width=0.32\linewidth,trim=0 0 0 17,clip]{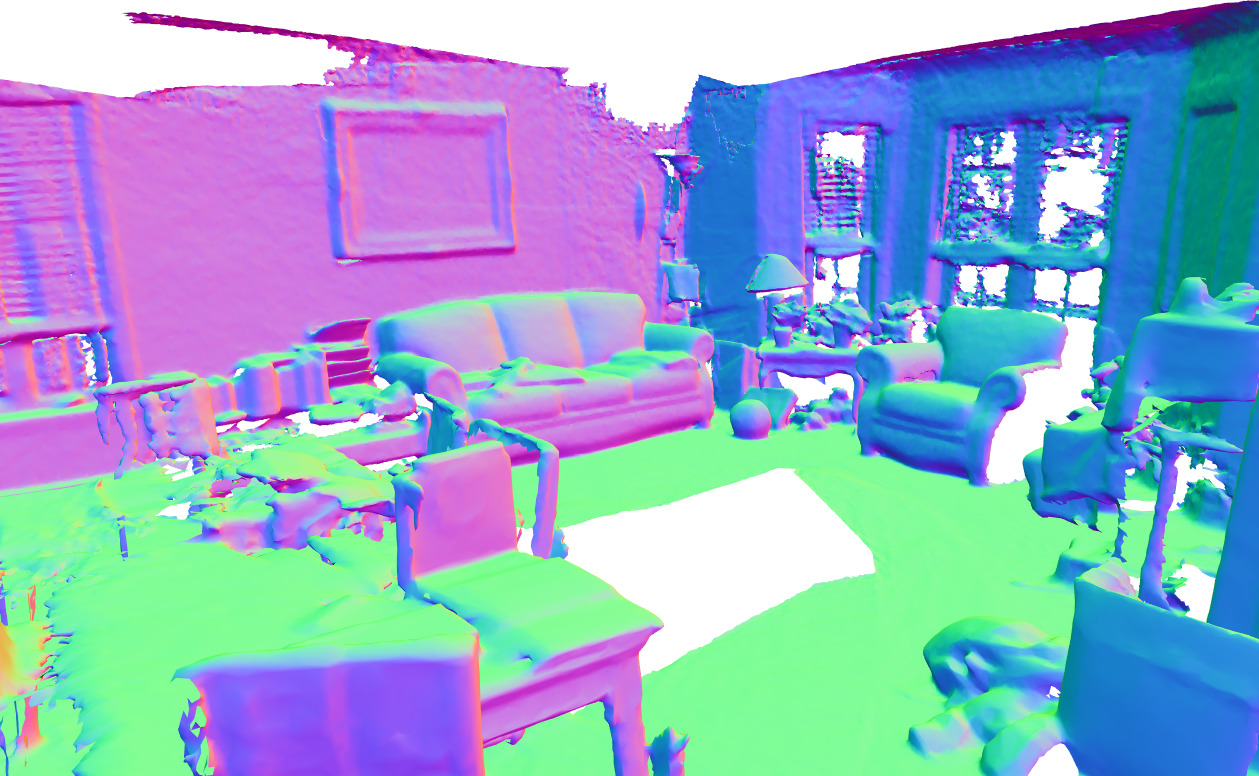}
  };
  \node[right = 0ex of 32] (33) {
    \includegraphics[width=0.32\linewidth,trim=0 28 0 93,clip]{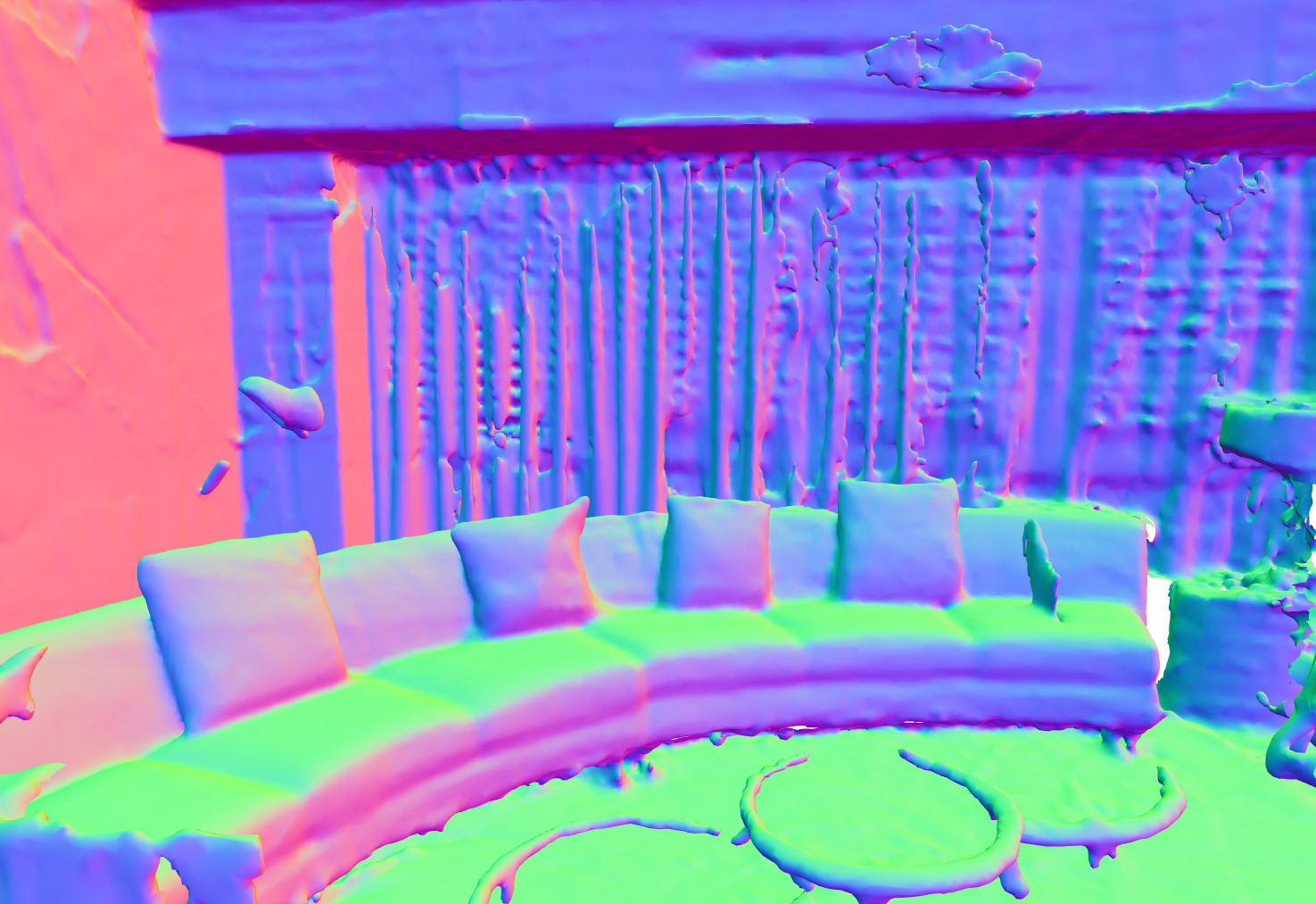}
  };
    
  \node[below = 0ex of 31] (41) {
    \scalebox{1}[-1]{\includegraphics[width=0.32\linewidth,trim=130 0 0 0,clip]{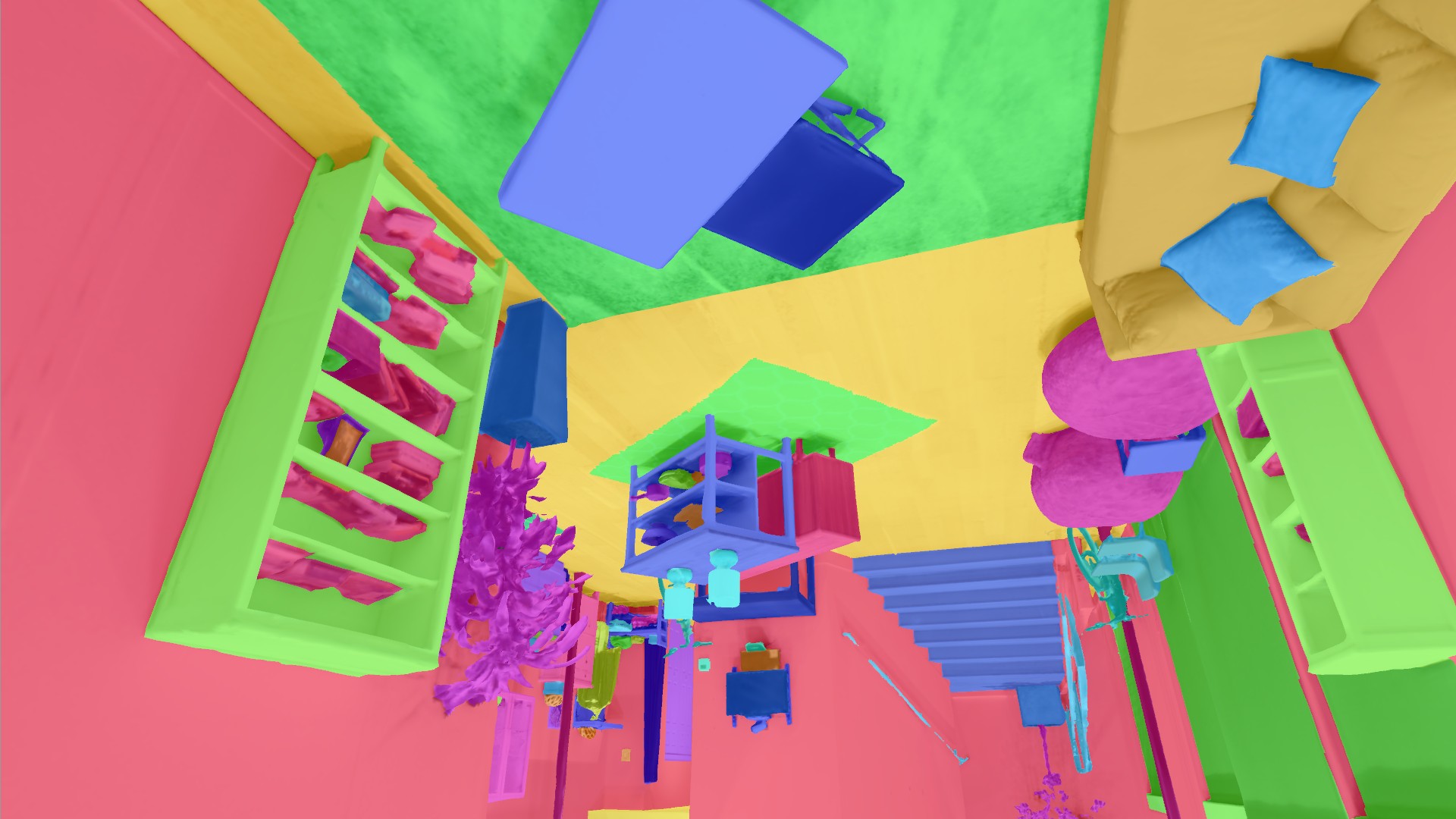}}
  };
  \node[right = 0ex of 41] (42) {
    \includegraphics[width=0.32\linewidth,trim=0 0 0 17,clip]{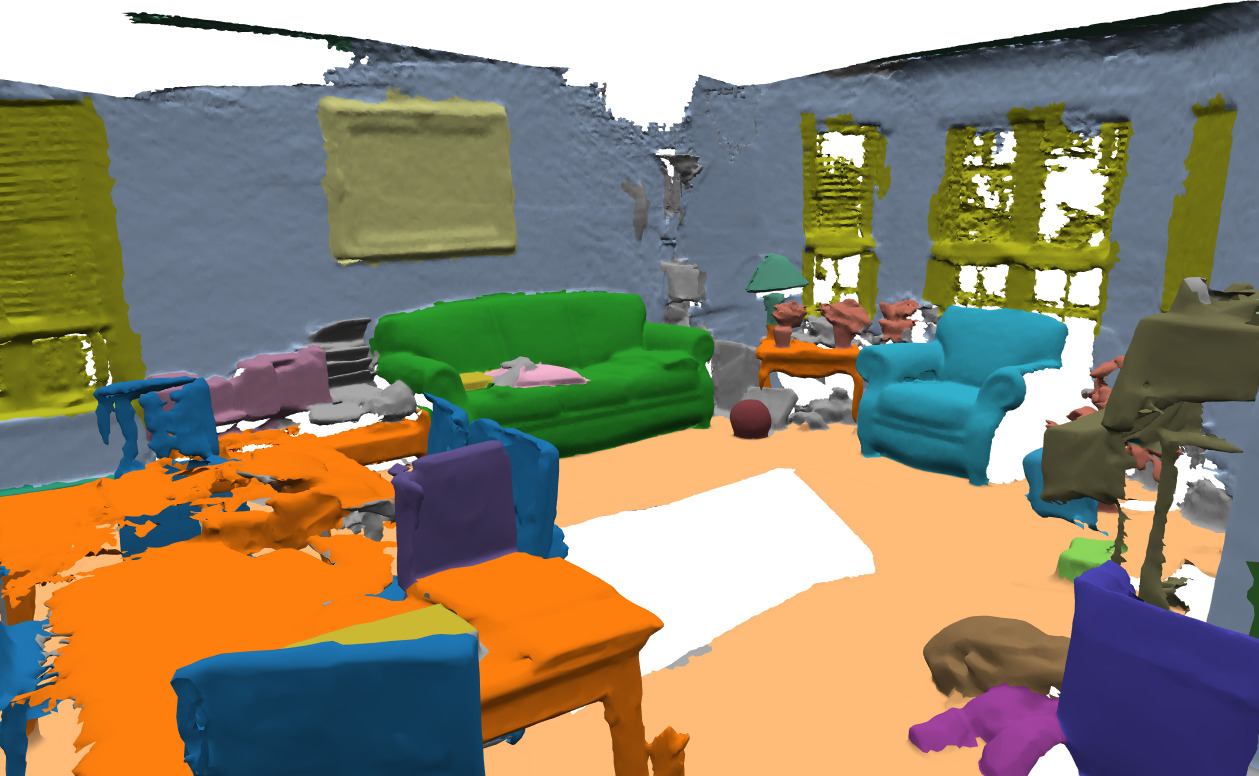}
  };
  \node[right = 0ex of 42] (43) {
    \includegraphics[width=0.32\linewidth,trim=0 28 0 93,clip]{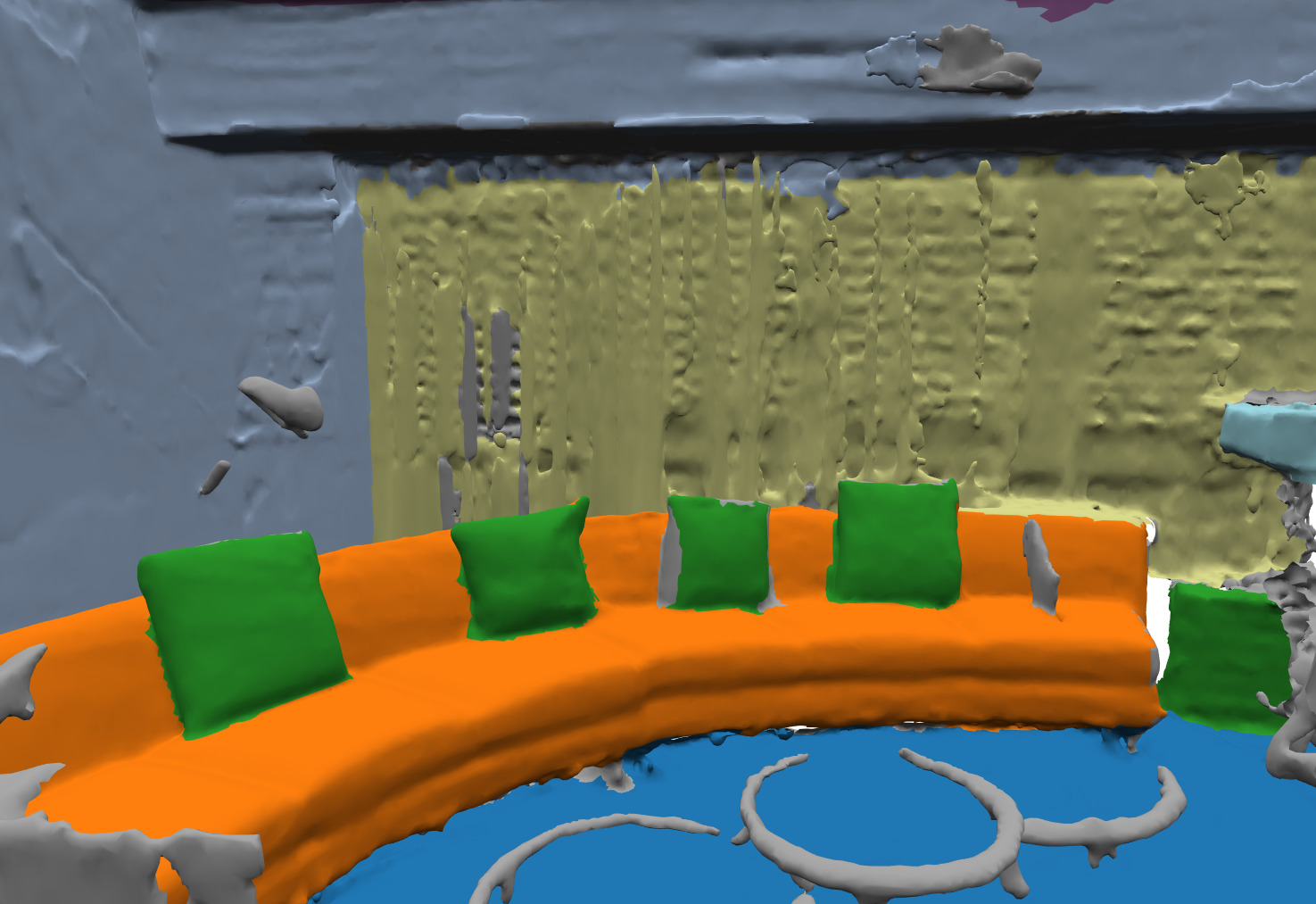}
  };
    
  \node[below = 0ex of 41] (51) {
    \scalebox{1}[-1]{\includegraphics[width=0.32\linewidth,trim=130 0 0 0,clip]{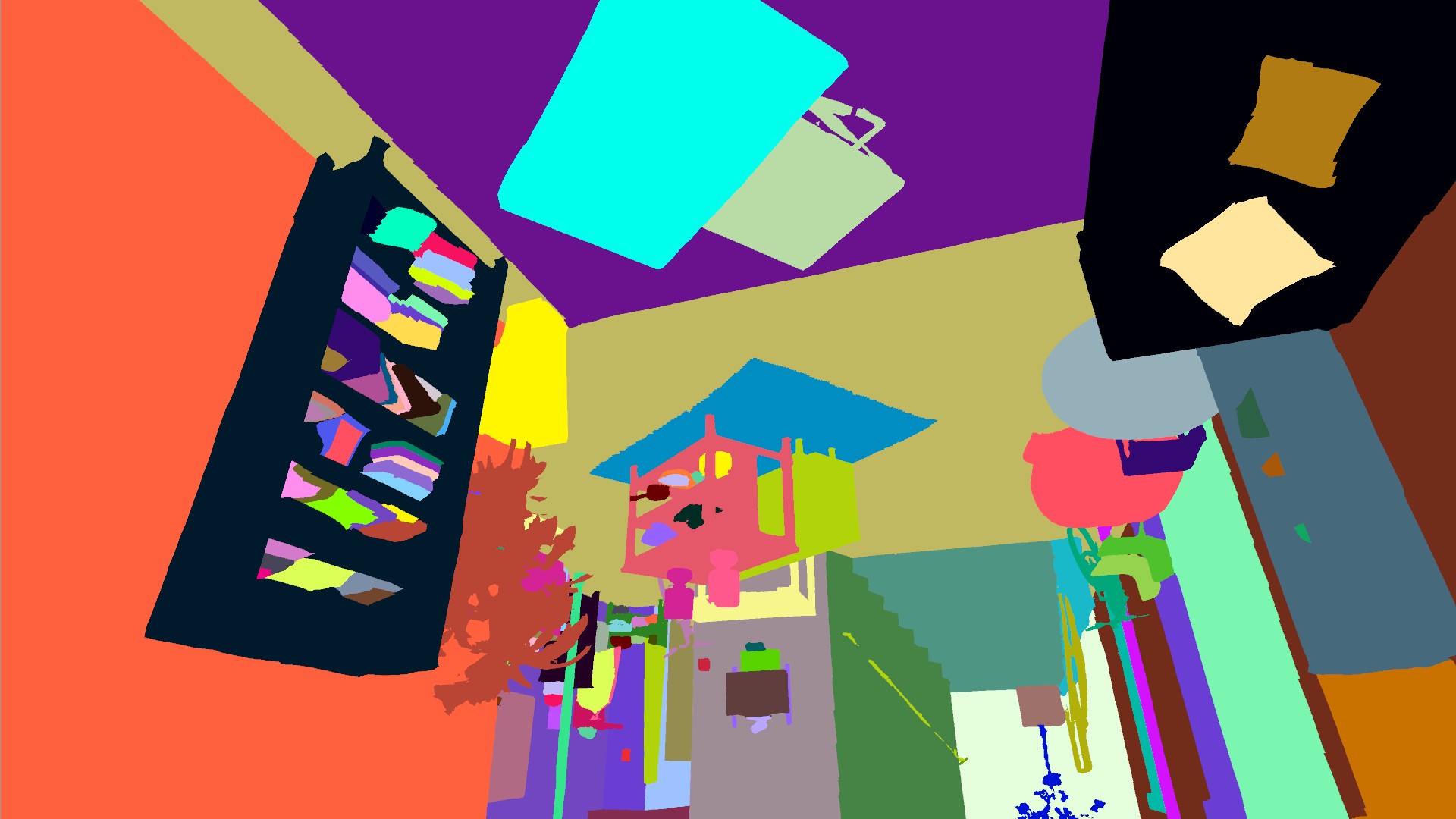}}
  };
  \node[right = 0ex of 51] (52) {
    \includegraphics[width=0.32\linewidth,trim=0 0 0 17,clip]{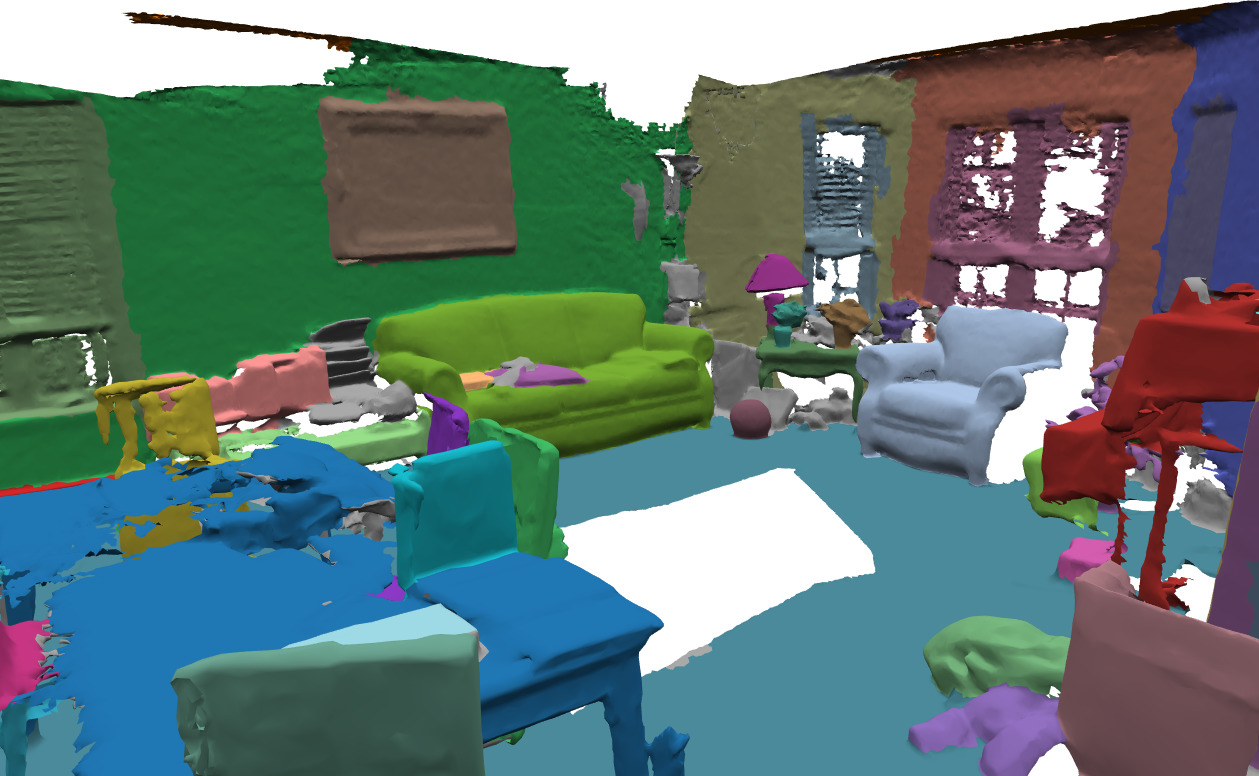}
  };
  \node[right = 0ex of 52] (53) {
    \includegraphics[width=0.32\linewidth,trim=0 28 0 93,clip]{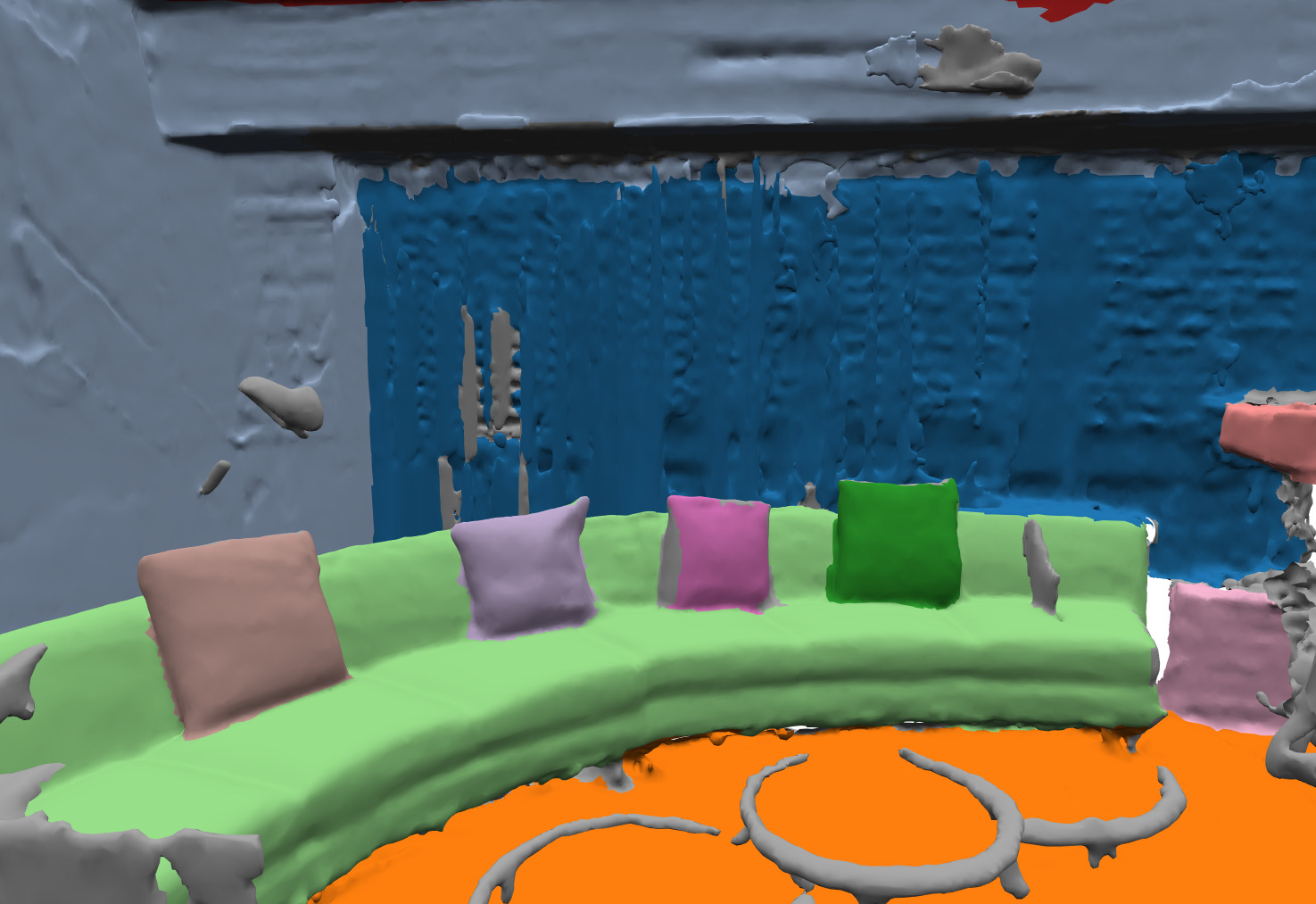}
  };

  \node[below = 0ex of 51] {Replica};
  \node[below = 0ex of 52] {ScanNet};
  \node[below = 0ex of 53] {Matterport 3D};

  \node[left = 0ex of 11,rotate=90,anchor=south] {RGB};
  \node[left = 0ex of 21,rotate=90,anchor=south] {geometry};
  \node[left = 0ex of 31,rotate=90,anchor=south] {normals};
  \node[left = 0ex of 41,rotate=90,anchor=south] {class seg.};
  \node[left = 0ex of 51,rotate=90,anchor=south] {instance seg.};

\end{tikzpicture}
    
    \caption{Renderings from comparison datasets to give a qualitative
    comparison to the Replica dataset. Note that clean geometry is important to
    allow rendering clean semantic class and instance segmentation. Geometry and texturing artifacts are noticeable in both Matterport 3D and ScanNet. Additionally ScanNet scans show a lot of missing surfaces and often do not capture the full room.}
    \label{fig:renderComparison}
\end{figure*}

\section{Related Work}

Existing 3D datasets can be classified broadly into two categories: (1) human-generated  synthetic scenes based on CAD models and (2) reconstructions of real environments. They vary in semantic and visual realism.

\subsection{Synthetic Scenes} 
%
SUNCG~\cite{Song2017} is a large dataset of synthetic indoor environments. However, the
scenes lack realistic appearances and are often semantically overly simplistic.
%
SceneNet~\cite{handa1511scenenet} is a synthetic dataset with 57 scenes and 3,699 object instances which can be automatically varied by sampling objects of the same class and similar size to replace the base objects in the 57 scenes.
%
The Stanford Scenes~\cite{2012-scenesynth} dataset consists of 130 scenes with 1,723 object instances.
On the smaller scale with only 16 scenes but with more realistic appearance is the RobotriX dataset~\cite{garcia2018robotrix}.
%
The InteriorNet~\cite{InteriorNet18} dataset consists of 22$M$ interior environments created from 1$M$ CAD assets. The dataset comes with 20$M$ images rendered out from the environments for SLAM benchmarking and machine learning. 
While newer synthetic datasets like InteriorNet are becoming more and more realistic, they still are not capturing real spaces with all their imperfections due to use, clutter and semantic variety.  

\subsection{Real Scenes}
There exists multiple datasets of 3D reconstructions of rooms and houses that
capture semantically realistic scenes as shown in the overview
Table~\ref{tab:related}. 
Based on Matterport's indoor scanning system there is the Matterport3D dataset~\cite{Chang2017}, the Gibson dataset~\cite{Xia2018}, and the Stanford 2D-3D-S dataset~\cite{armenicvpr16}, some of which capture hundreds of scenes.
These scales are impressive for reconstruction-based 3D scene datasets as it takes effort to collect, process, clean up and semantically annotate real data.
The visual quality of the Matterport-scanner-based datasets is more realistic
than SUNCG but geometry artifacts and lighting problems exist throughout the datasets, as shown in Fig.~\ref{fig:renderComparison}.

The original Matterport3D~\cite{Chang2017} dataset consists of 90 houses with 2,056 rooms and 50,811 object instances from 40 semantic classes. Semantic annotation was performed based on a 3D Felsenszwalb pre-segmentation~\cite{felzenszwalb2004efficient}. This means the resolution and accuracy of the semantic annotation is constrained to the segments extracted by the Felsenszwalb algorithm, which we found to be prone to inaccuracy on boundaries between objects.
%
The Stanford 2D-3D-S dataset~\cite{armenicvpr16} contains 6 large-scale reconstructions with a total of 270 rooms. It is annotated with 13 object classes and 11 scene categories. The exact method of semantic annotation is not described except that it is done in 3D. 
%
The Gibson dataset~\cite{Xia2018} contains 572 buildings and includes the two aforementioned datasets. Only the meshes from the Matterport3D and the Stanford 2D-3D-S dataset contain semantic segmentations.

Beyond Matterport-scanner-based reconstructions, there is the ScanNet~\cite{dai2017scannet} dataset which was obtained by scanning scenes with an iPad-based RGB-D camera system.
It contains 1,513 scenes with more than 19 scene types and a flexible yet unspecified number of semantic classes. Mapping of the semantic classes to NYU~v2, ModelNet, ShapeNet and WordNet exists. Semantic annotation was performed based on a Felsenszwalb segmentation with the same downside of inaccurate segmentation boundaries as described previously.

Table~\ref{tab:related} shows that while this initial release of Replica is a smaller dataset, its
reconstructions have high color, geometry, and semantic resolution.
Additionally, the Replica dataset introduces HDR textures and renderable reflectors.





\section{Dataset Creation}

\begin{figure}
    \centering
    \includegraphics[width=\linewidth]{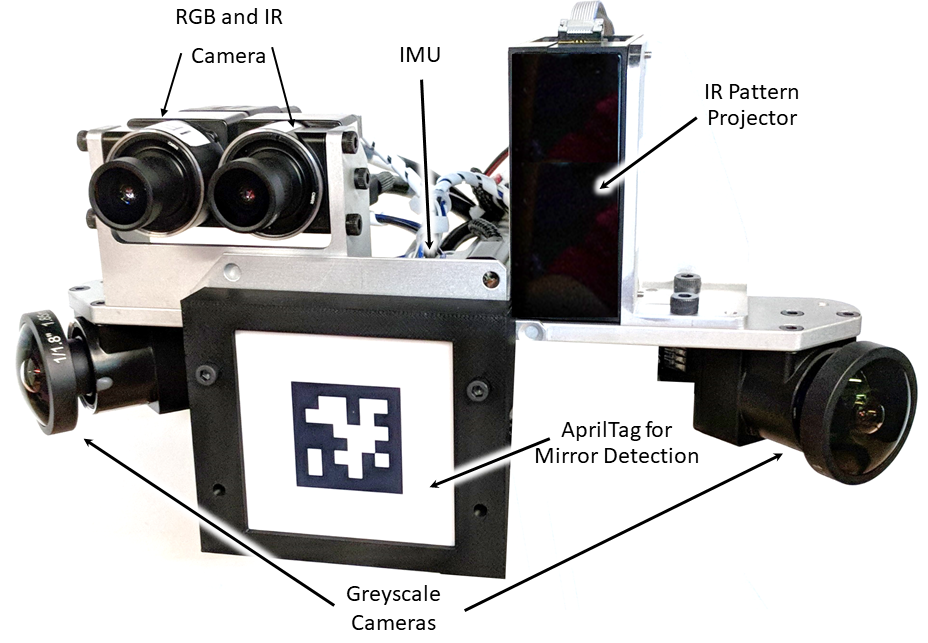}
    \caption{The data collection rig used to capture the raw data used to build Replica.}
    \label{fig:rig}
\end{figure}

\begin{figure*}
    \centering
    \includegraphics[width=0.49\linewidth]{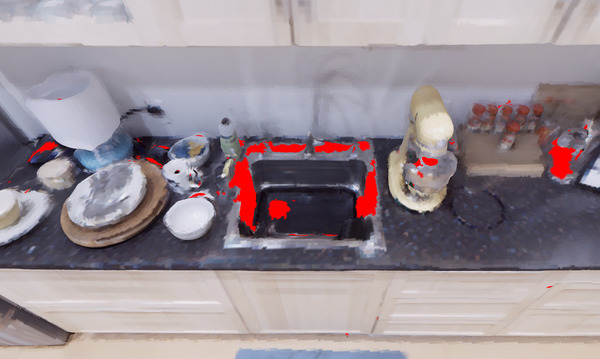}
    \includegraphics[width=0.49\linewidth]{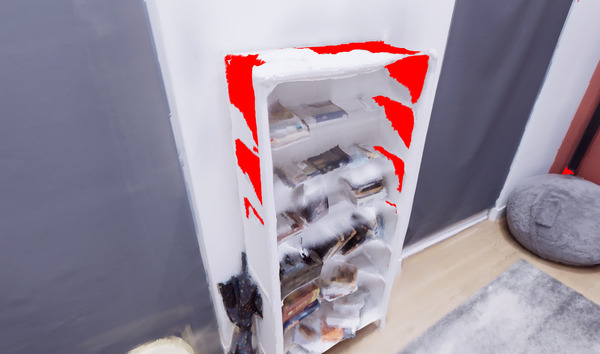}
    
    \includegraphics[width=0.49\linewidth]{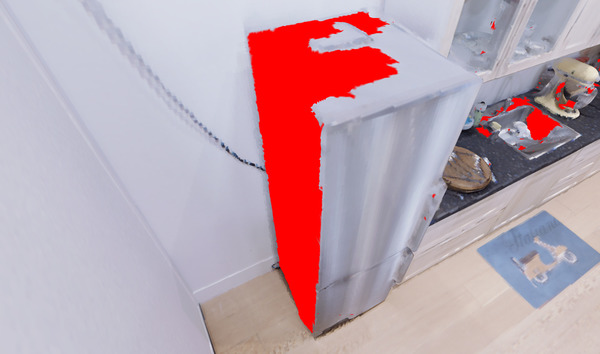}
    \includegraphics[width=0.49\linewidth]{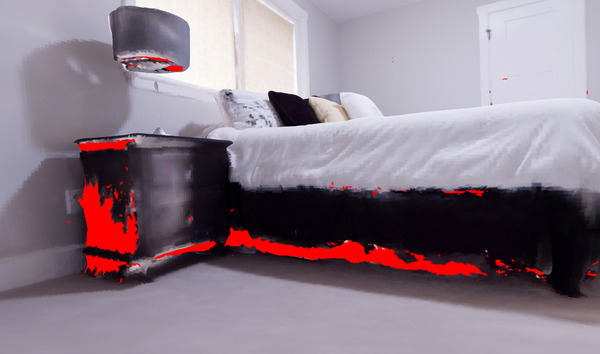}
    \includegraphics[width=0.49\linewidth]{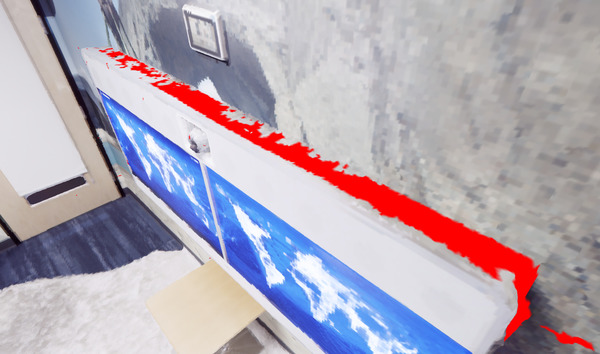}
    \includegraphics[width=0.49\linewidth]{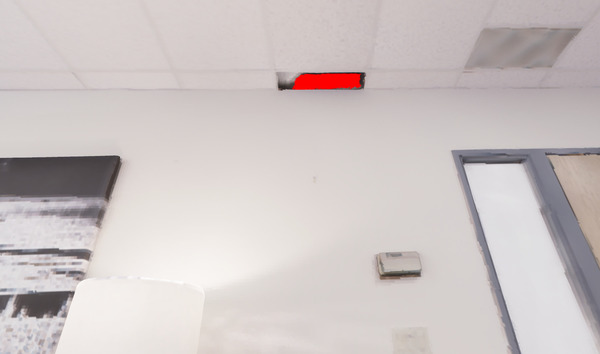}

    \caption{Example of holes filled with the mesh fix-up tool. Filled holes are marked red. In these images the textures have been tonemapped to a low dynamic range to facilitate easier human interpretation for manual touch up.}
    \label{fig:holeFillingExample}
\end{figure*}

To create the Replica reconstructions, we use a custom built RGB-D capture rig with an IR projector depicted in Fig.~\ref{fig:rig}. It collects time-aligned raw IMU, RGB, IR and wide-angle greyscale sensor data. 
The wide-angle greyscale video data together with the IMU data is used by an
in-house SLAM system, similar to state-of-the-art systems
like~\cite{engel2017direct,mur2015orb}, to provide 6 degree of freedom (DoF) poses.
We compute raw depth from the IR video stream given the IR structured light pattern projected from the rig. Given the 6~DoF poses from the SLAM system, depth images are fused into a truncated signed distance function (TSDF) akin to KinectFusion~\cite{newcombe2011kinectfusion}.
Meshes are extracted using the standard Marching Cubes~\cite{Lorensen1987MC} algorithm, simplified via Instant Meshes~\cite{Jakob2015Instant} and textured with a PTex-like system~\cite{burley2008ptex}. Finally, we extract mirrors and reflective surfaces~\cite{whelan2018reconstructing}.

HDR textures are obtained by cycling the exposure times of the RGB texture camera and, using the 6~DoF SLAM poses, fusing the measured radiance per texel into 16~bit floating point RGB values. This approach yields an overall dynamic range of about 85,000:1 which corresponds to more than 16~f-stops as opposed to the standard vertex mesh colors and textures of the other datasets which are encoded as 8~bit RGB values.

\subsection{Mesh and Reflector Fixing}
To ensure the highest quality 3D meshes, we manually fix planar reflective surfaces and small holes where surfaces were not sufficiently captured during scanning. 
Reflective surfaces are defined as planar polygons and can be annotated in our custom built software tool by specifying the boundary of the reflector on the mesh.
For hole filling we first automatically detect holes by searching for boundary edges that form closed cycles and hence constitute holes. 
A human annotator can then use our tool to select a hole and automatically fill it using the approach described by Liepa~\cite{liepa2003filling}. Specifically, we use CGAL~\cite{cgalpmp2019} to triangulate the hole boundary to generate an initial patch, then refine and smooth the patch. Examples of patched holes are shown in Fig.~\ref{fig:holeFillingExample}.

\subsection{Semantic Annotation}

%
Semantic annotation is performed in two steps. First, we render a set of images from the mesh such that all primitives of the mesh are observed at least once. These images are then annotated in parallel using a 2D instance-level masking tool. After 2D annotation, we fuse the 2D semantic annotations back onto the mesh using a voting scheme. The 3D annotations are then refined using a superpixel-like segmentation. This ensures that small holes in the initial fused segmentation are filled based on neighborhood information.
In the second step we review, refine and correct the fused segmentation using a 3D annotation tool that in effect allows painting on the 3D mesh.
This step ensures highest annotation quality since annotations can be refined down to the primitive level. 

As part of the semantic annotation we also annotate areas that need to be anonymized (i.e. blurred or pixelated) to ensure privacy. 

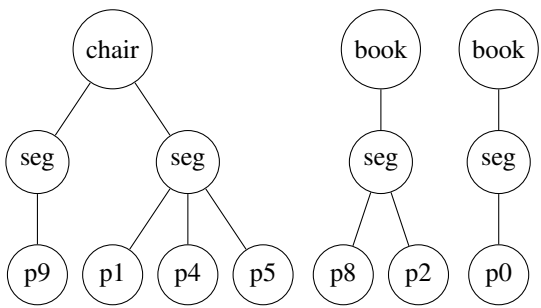
\begin{figure}
    \centering
\begin{tikzpicture}[
level distance=1.5cm,
  level 1/.style={sibling distance=2cm},
  level 2/.style={sibling distance=1.cm},
  every node/.style = {shape=circle,
    draw, align=center}]]
  \node (chair) {chair}
    child { node {seg} 
      child { node {p9} }
    }
    child { node {seg}
      child { node {p1} }
      child { node {p4}  }
      child { node (p5) {p5} }  
      };
  \node (book) [right = of chair, xshift=1.5cm] {book}
    child { node {seg}
      child { node {p8} }
      child { node {p2}  }
      };
  \node[right = of book, xshift=-0.5cm] {book}
    child { node {seg}
      child { node {p0} }
      };
\end{tikzpicture}
    \caption{In the proposed segmentation forest data structure, the root of each tree indicates the semantic object instance. 
		The mesh primitives from the leaf nodes (denoted ``p'') are connected into segmentation nodes (denoted ``seg'') one level below the roots.}
    \label{fig:segmentationForest}
\end{figure}

We represent the semantic annotation as a multi-tree or forest data structure which we call a segmentation forest: At the bottom level are the individual primitives of the mesh. The next level connects primitives into larger segments. At the root level these segments are connected into semantic object entities. Figure~\ref{fig:segmentationForest} shows a simple example comprised of a chair and two book instances. As can be seen, the segmentation forest data structure represents an instance segmentation of the scene where each tree in the semantic annotation forest corresponds to a semantic instance. 
A class segmentation is obtained by simply rendering all instances of the same class in the same color. 
The segmentation forest data structure is flexible in that it allows connecting semantic instances in a hierarchical way. Rendering at different levels of the forest leads to different segmentations of the scene.

\section{Dataset Description \label{sec:dataset}}

\begin{figure*} 
\centering
\begin{subfigure}[t]{0.32\linewidth}
\includegraphics[width=\linewidth]{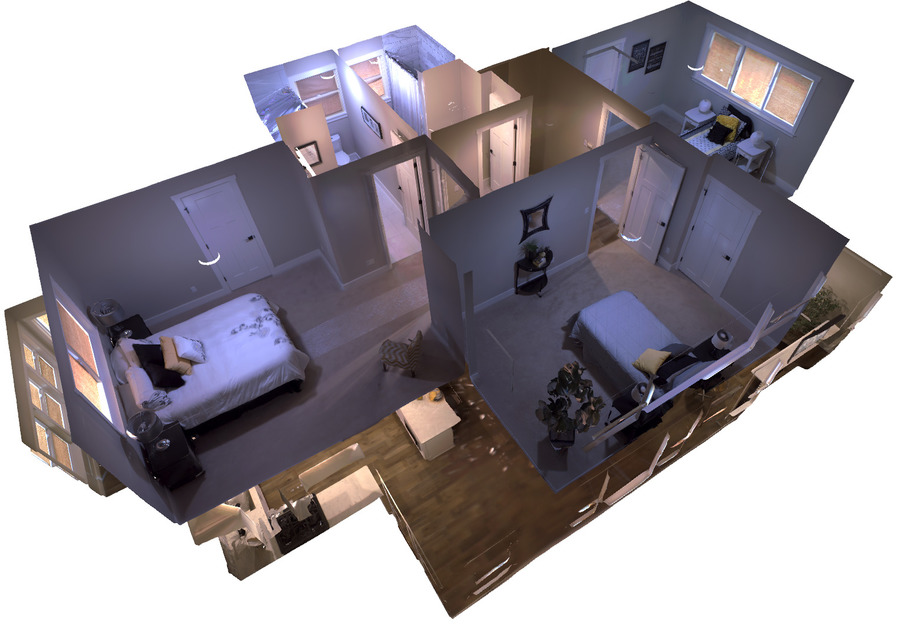}
\caption{apartment 0}
\end{subfigure}
\begin{subfigure}[t]{0.32\linewidth}
\includegraphics[width=\linewidth]{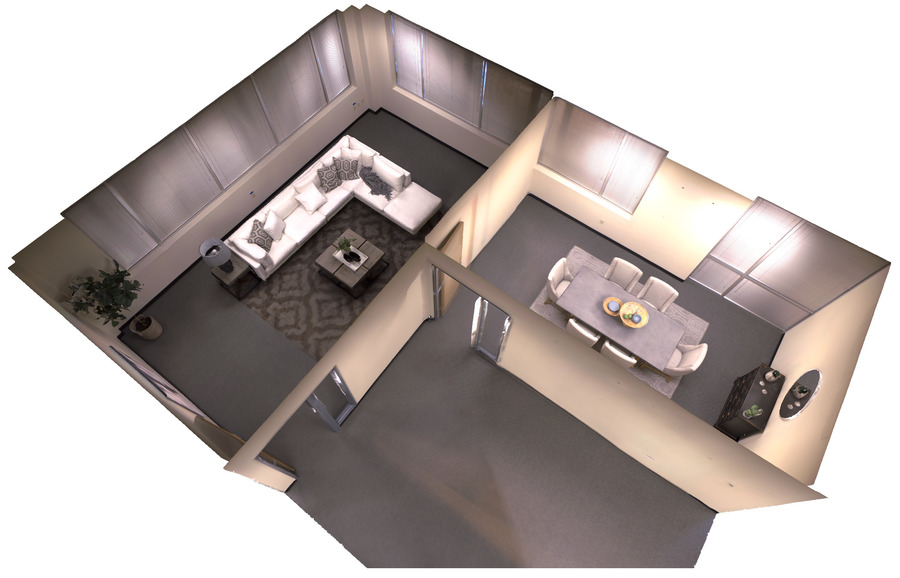}
\caption{apartment 1}
\end{subfigure}
\begin{subfigure}[t]{0.32\linewidth}
\includegraphics[width=\linewidth]{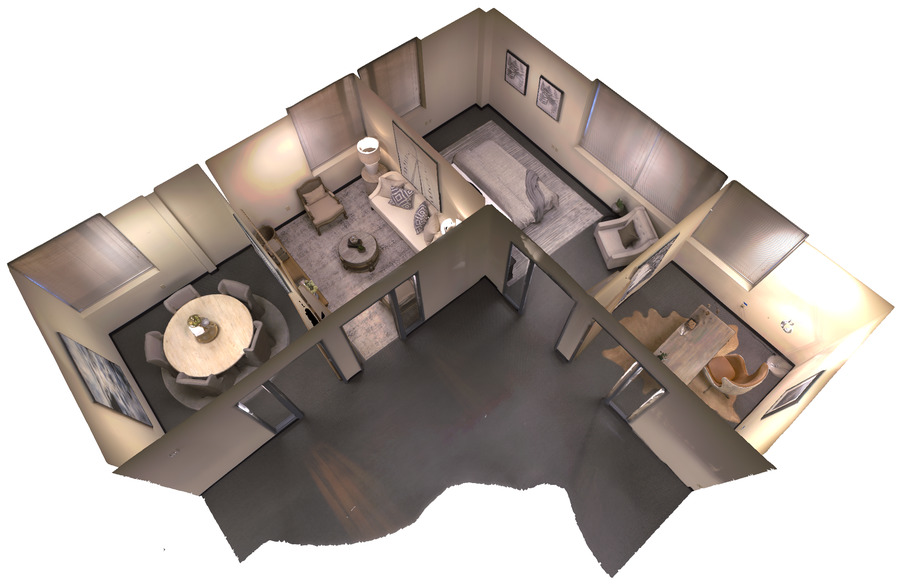}
\caption{apartment 2}
\end{subfigure}
\begin{subfigure}[t]{0.32\linewidth}
\centering
\includegraphics[width=0.9\linewidth]{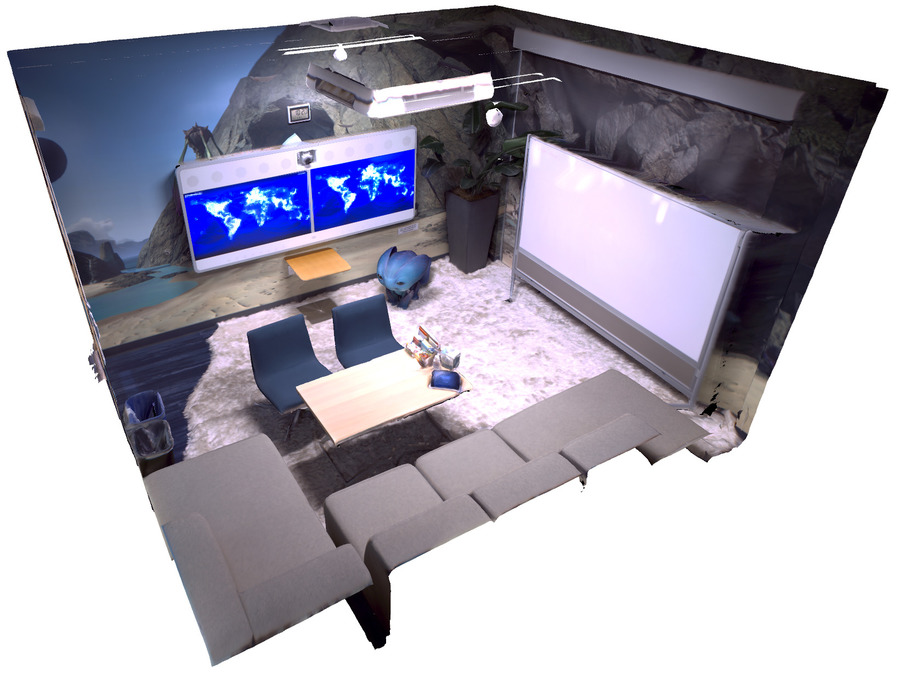}
\caption{office 0}
\end{subfigure}
\begin{subfigure}[t]{0.32\linewidth}
\includegraphics[width=\linewidth]{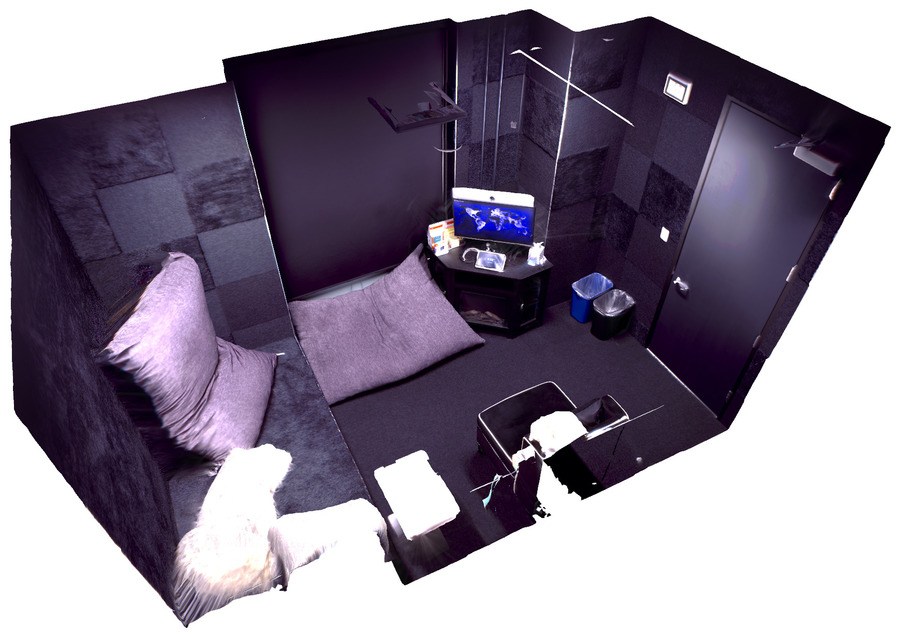}
\caption{office 1}
\end{subfigure}
\begin{subfigure}[t]{0.32\linewidth}
\includegraphics[width=\linewidth]{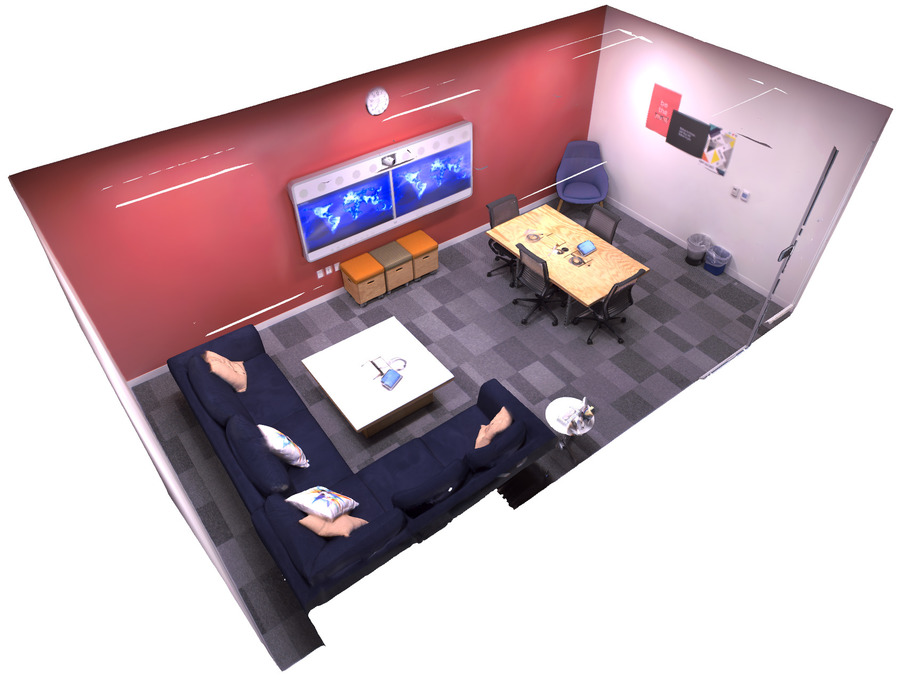}
\caption{office 2}
\end{subfigure}
\begin{subfigure}[t]{0.32\linewidth}
\includegraphics[width=\linewidth]{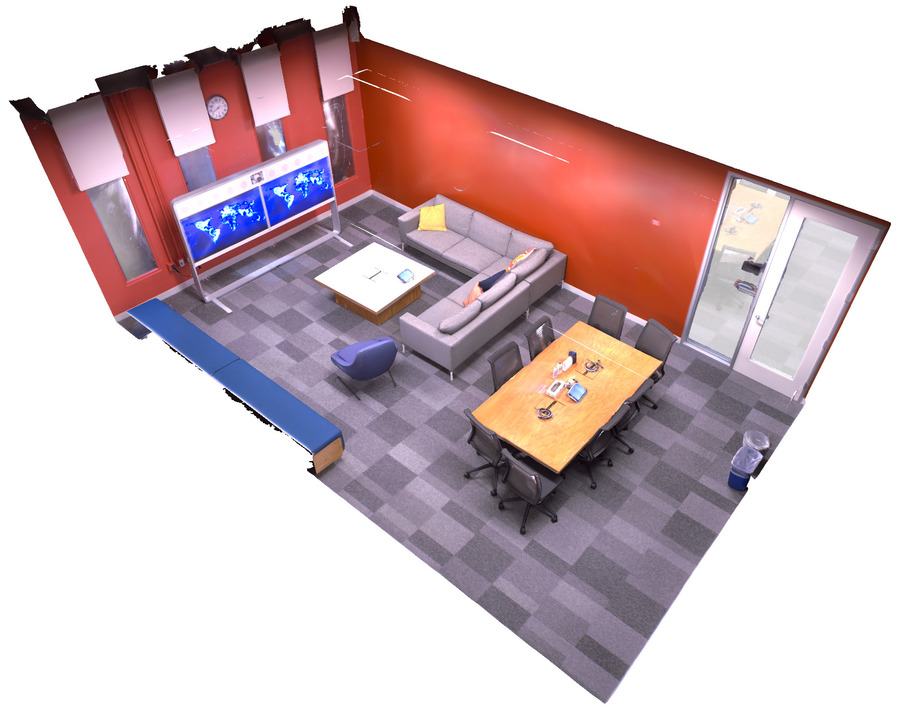}
\caption{office 3}
\end{subfigure}
\begin{subfigure}[t]{0.32\linewidth}
\includegraphics[width=\linewidth]{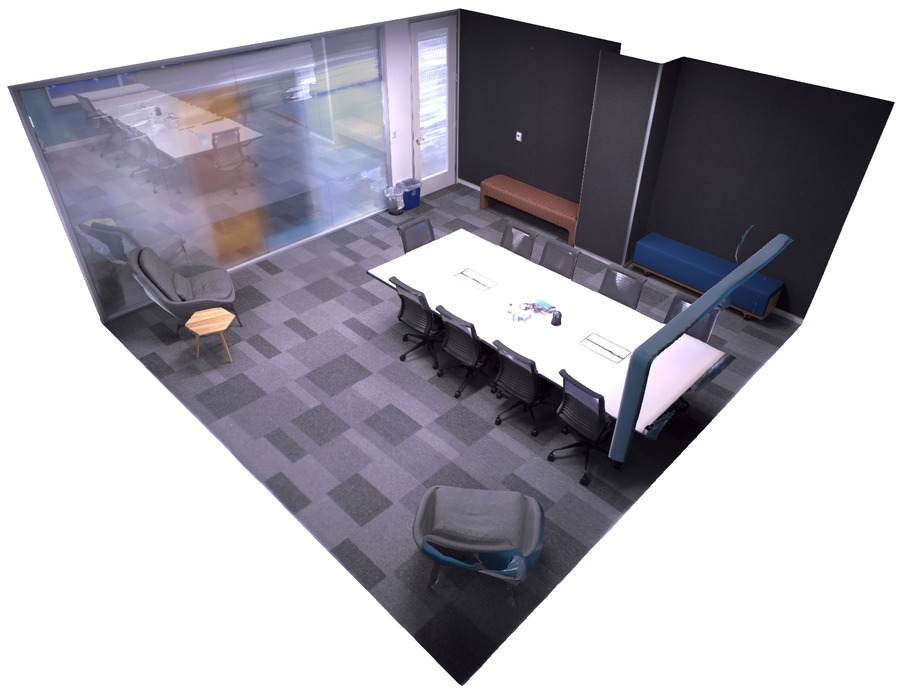}
\caption{office 4}
\end{subfigure}
\begin{subfigure}[t]{0.32\linewidth}
\includegraphics[width=\linewidth]{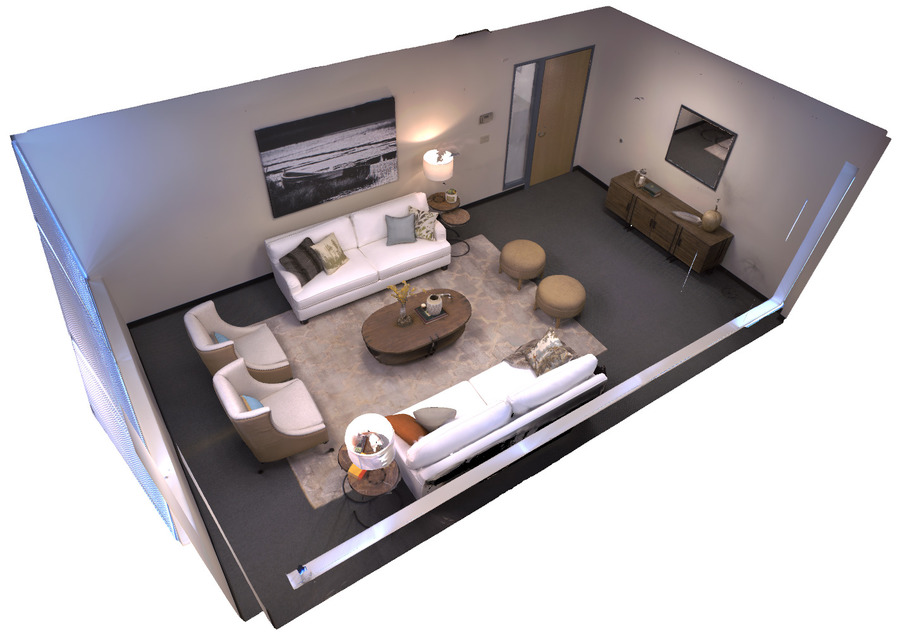}
\caption{room 0}
\end{subfigure}
\begin{subfigure}[t]{0.32\linewidth}
\includegraphics[width=\linewidth]{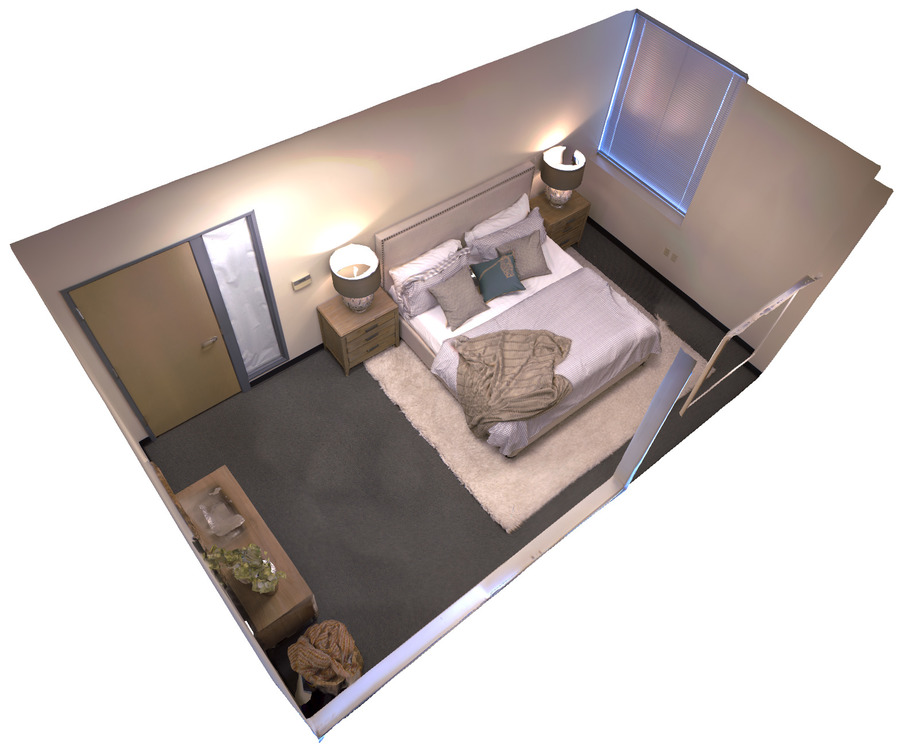}
\caption{room 1}
\end{subfigure}
\begin{subfigure}[t]{0.32\linewidth}
\includegraphics[width=\linewidth]{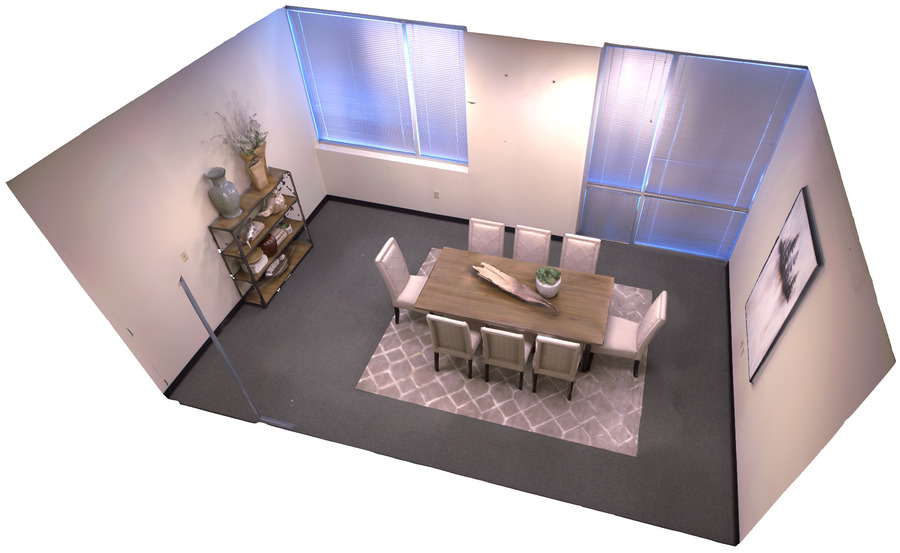}
\caption{room 2}
\end{subfigure}
\begin{subfigure}[t]{0.32\linewidth}
\includegraphics[width=\linewidth]{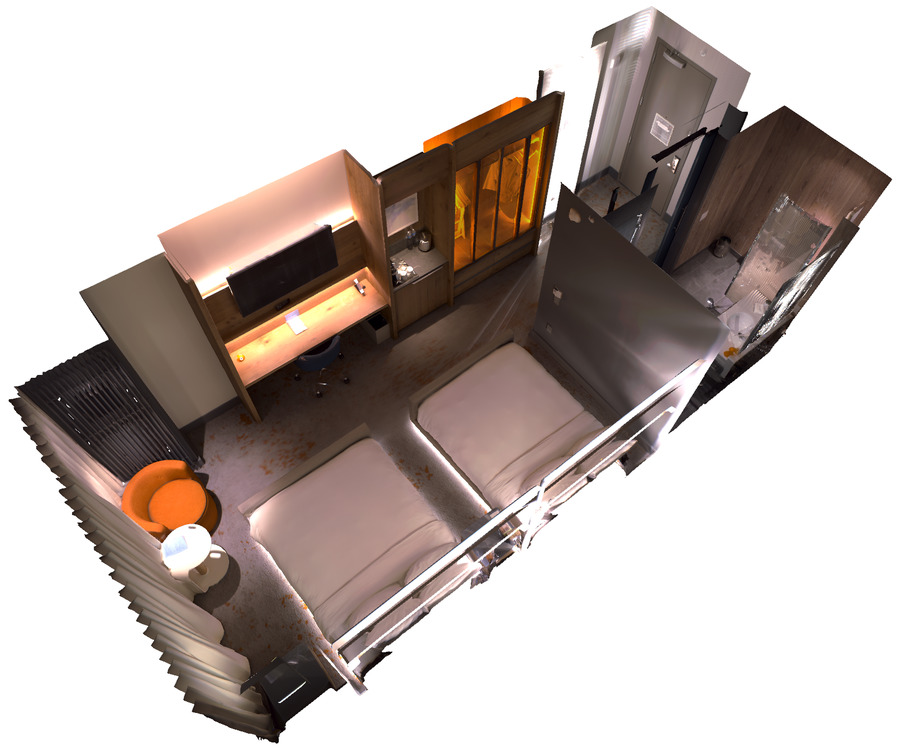}
\caption{hotel 0}
\end{subfigure}

\caption{The Replica dataset contains a variety of \nMeshWoFRLApt{} semantically different reconstructions.}
    \label{fig:ReplicaScenes}
\end{figure*} 

\begin{figure*}
\centering
\begin{subfigure}[t]{0.32\linewidth}
\includegraphics[width=\linewidth]{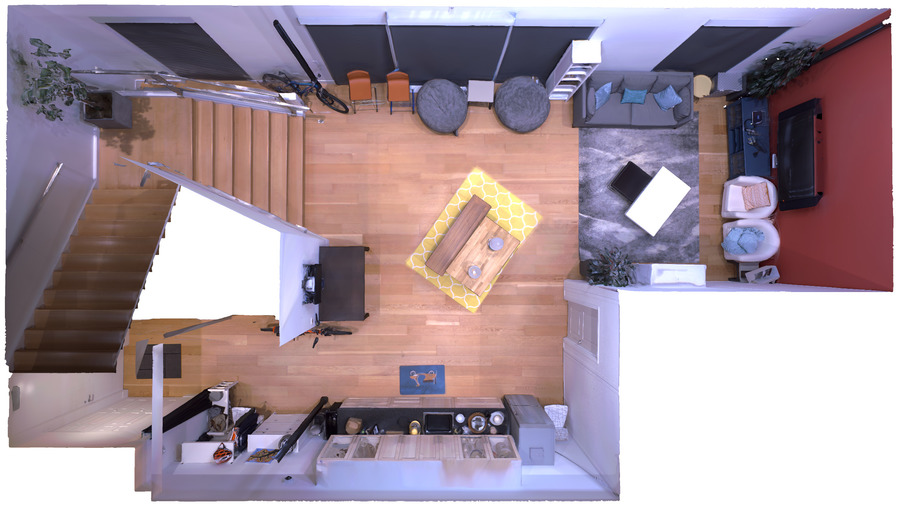}
\caption{FRL apartment 0}
\end{subfigure}
\begin{subfigure}[t]{0.32\linewidth}
\includegraphics[width=\linewidth]{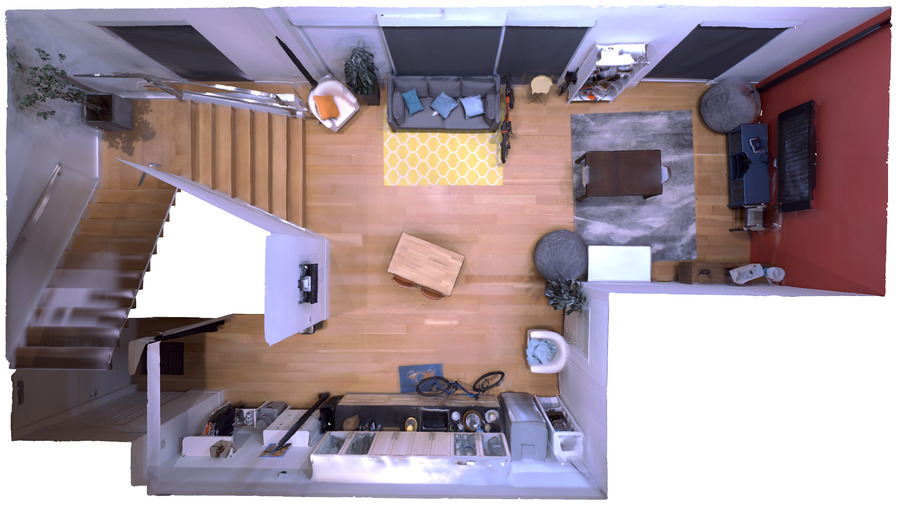}
\caption{FRL apartment 1}
\end{subfigure}
\begin{subfigure}[t]{0.32\linewidth}
\includegraphics[width=\linewidth]{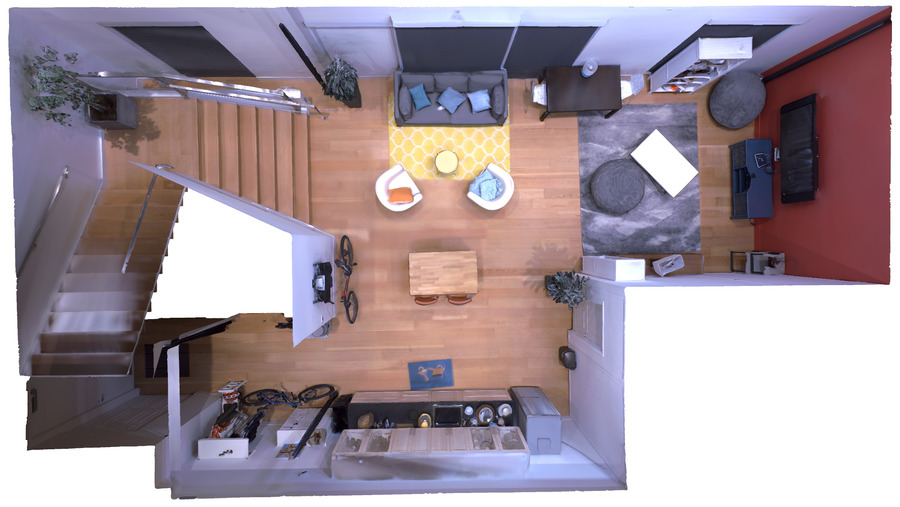}
\caption{FRL apartment 2}
\end{subfigure}
\begin{subfigure}[t]{0.32\linewidth}
\includegraphics[width=\linewidth]{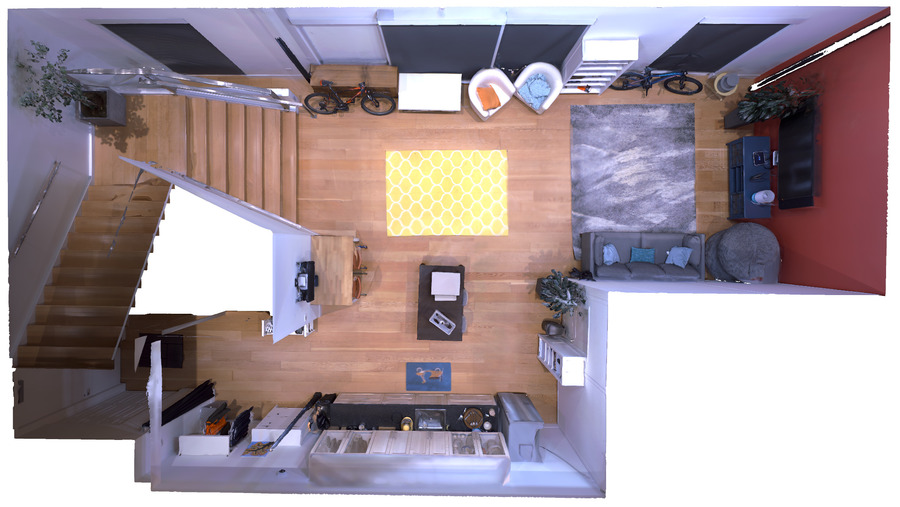}
\caption{FRL apartment 3}
\end{subfigure}
\begin{subfigure}[t]{0.32\linewidth}
\includegraphics[width=\linewidth]{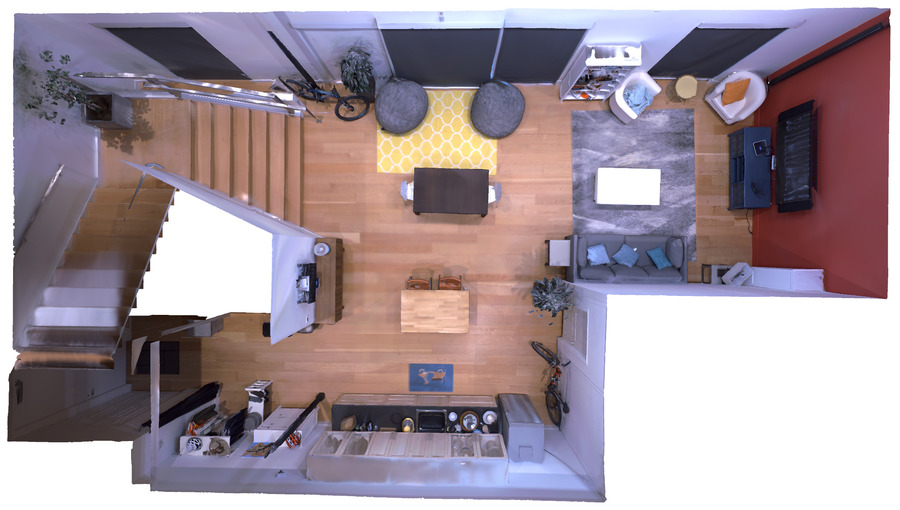}
\caption{FRL apartment 4}
\end{subfigure}
\begin{subfigure}[t]{0.32\linewidth}
\includegraphics[width=\linewidth]{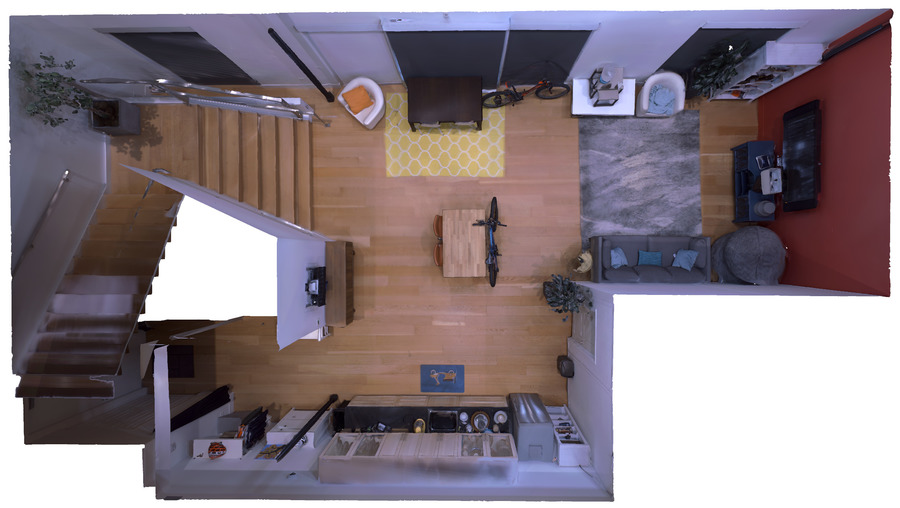}
\caption{FRL apartment 5}
\end{subfigure}

\caption{The Replica dataset contains a set of \nFRLApt{} scenes of the FRL apartment with the contents rearranged mimicking the same scene at different points in time.}
    \label{fig:ReplicaScenesFRL}
\end{figure*} 

\begin{figure*}
    \centering
      \includegraphics[height=0.26\paperheight]{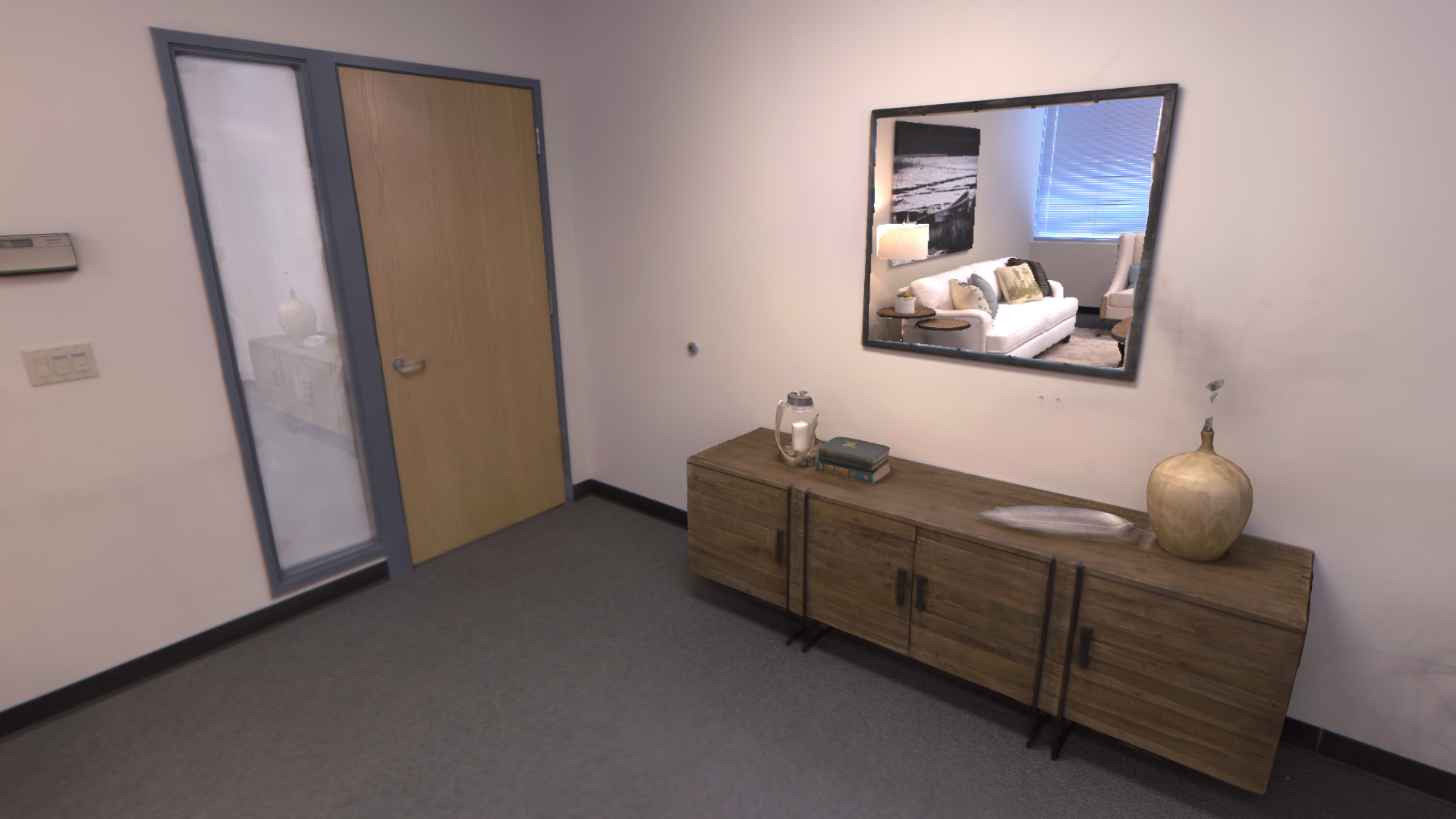}\\[5pt]
      \includegraphics[height=0.26\paperheight]{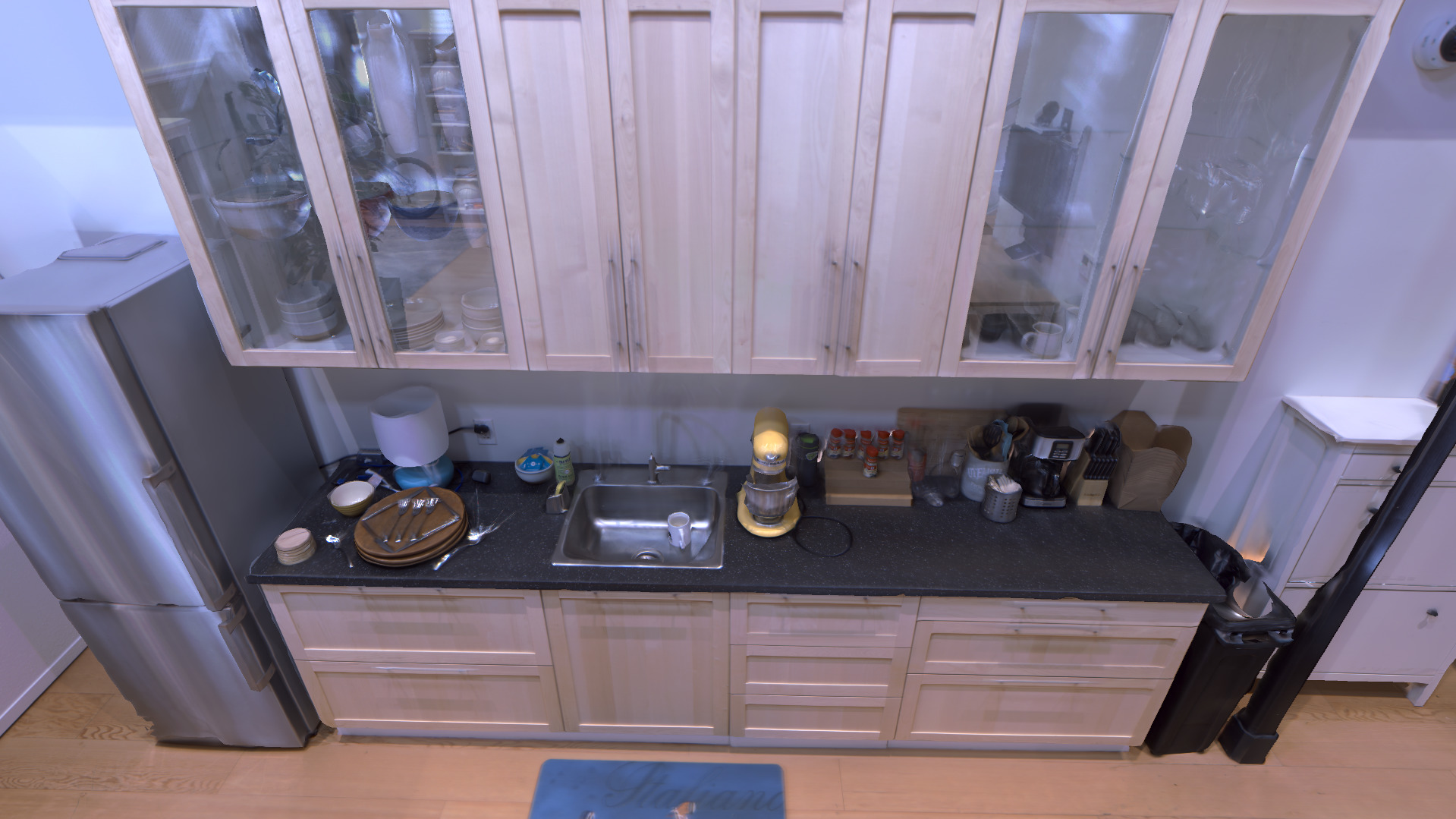}\\[5pt]
      \includegraphics[height=0.26\paperheight]{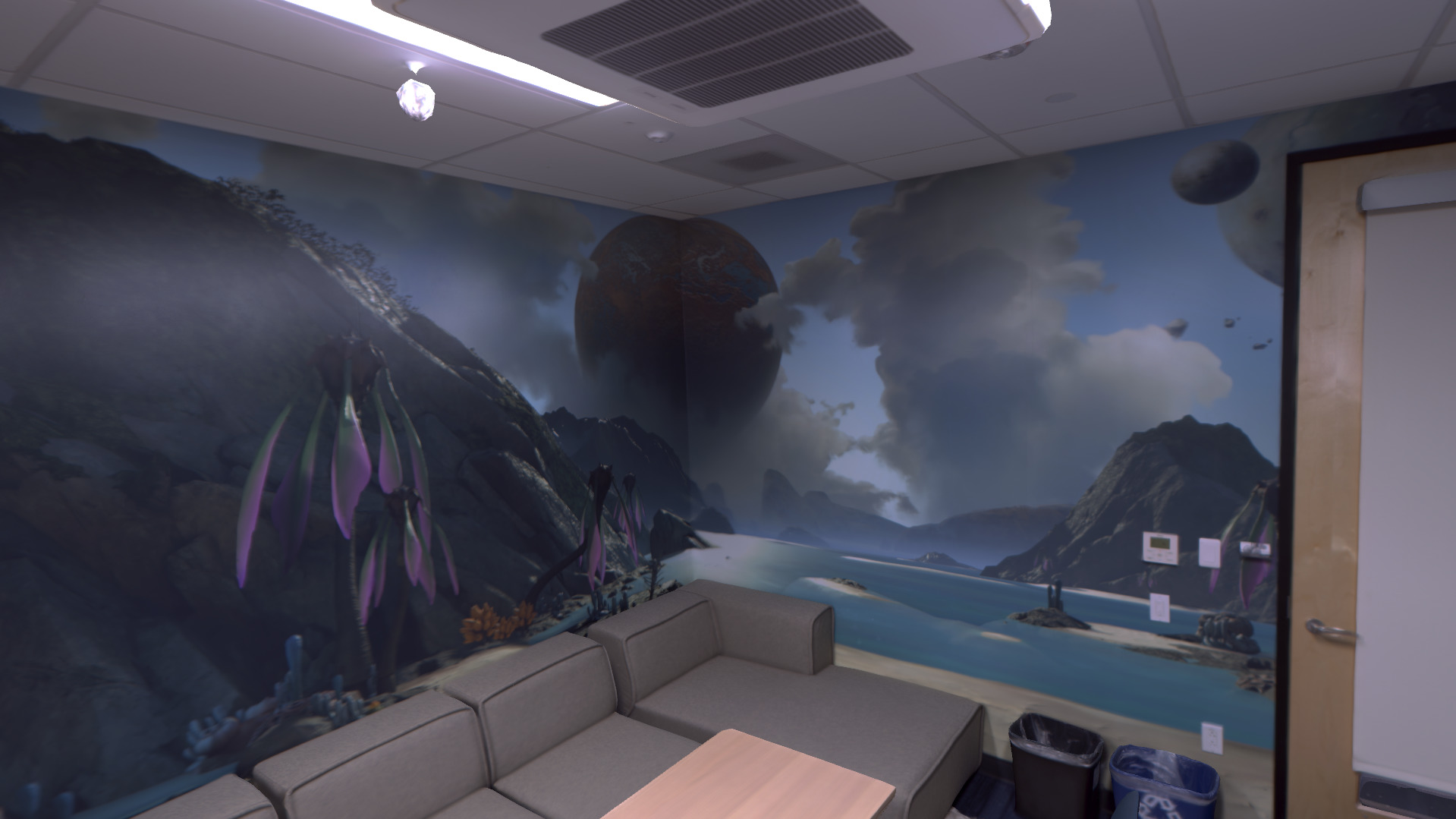}

    \caption{Example renderings from the Replica dataset showing glass and mirror
    reflectors as well as high resolution textures.}
    \label{fig:rendersReplica}
\end{figure*}

The Replica dataset together with a minimal SDK are published at the following github repository: \href{url}{https://github.com/facebookresearch/Replica-Dataset}.

As shown in Fig.~\ref{fig:ReplicaScenes} and \ref{fig:ReplicaScenesFRL}, the Replica dataset contains \nMesh{}
different scenes: \nFRLApt{} different setups of the FRL apartment, \nOffice{} office rooms, a
2-floor house, \nAptMultiRoom{} multi-room apartment spaces, a hotel room, and \nAptRooms{} rooms of apartments. The scenes were selected with an
eye towards semantic variety of the environments as well as their scale. With
the \nFRLApt{} FRL apartment scenes with different setups we introduce a dataset of
scenes taken at different points in time of the same space.

Each Replica scene contains dense geometry, high resolution HDR textures,
reflectors and semantic class and instance annotation as shown for one of the
datasets in Fig.~\ref{fig:teaser}. Figure~\ref{fig:renderComparison} shows
renderings from the FRL Apartment dataset for the different modalities. Note the
high fidelity of the semantic annotations and the accuracy at borders.

As shown in Fig.~\ref{fig:rendersReplica} glass and mirror surface information
is contained in the Replica dataset and can be rendered for additional realism
and photometric accuracy.

In Fig.~\ref{fig:photometric} we show comparisons of the raw RGB image captured from the data collection rig next to a rendering of the scene from same pose. Qualitatively, it is hard to tell whether the left or right frames are the raw captures underscoring the realism of the Replica reconstructions. Small artifacts and the fact that there is no motion blur give away that the right column shows the rendered images. Additionally, the foot of the operator is accidentally captured in the second example giving another hint that the left column contains the raw captured images.

\begin{figure*}
    \centering
    \includegraphics[width=\linewidth]{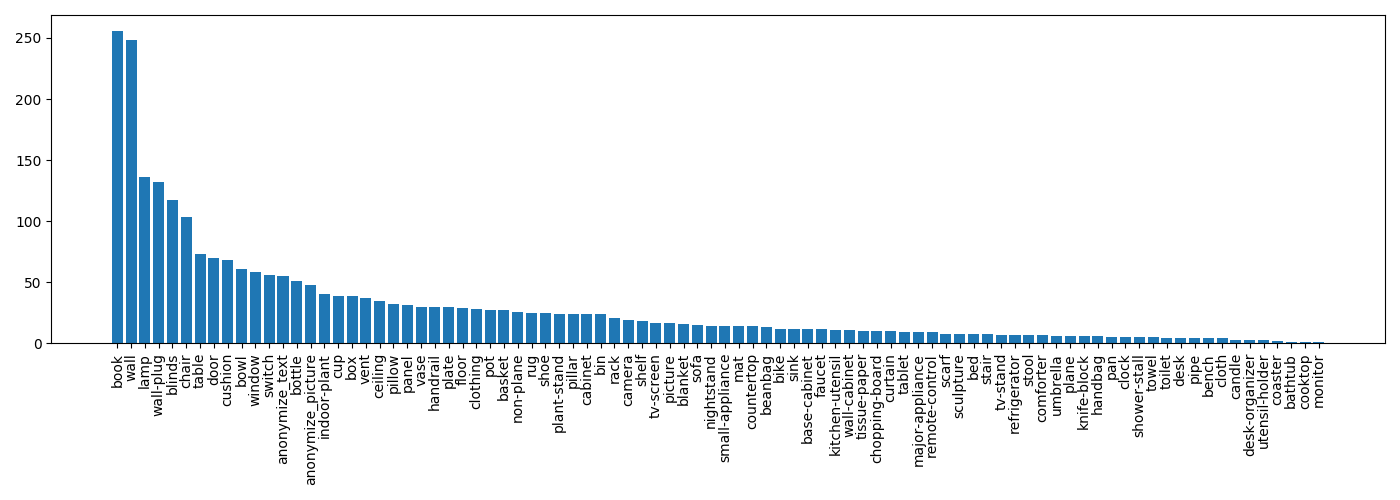}
    \caption{Histogram over the \nClass{} semantic classes contained in the dataset.}
    \label{fig:instanceHist}
\end{figure*}

Figure~\ref{fig:instanceHist} shows a histogram over semantic instances across
the dataset. The semantic classes were picked to capture the variety of objects
and surface classes in Replica. The figure shows that common structural elements
such as ``floor'', ``wall'', ``ceiling'' as well as various object types from
``chair'' to ``book'' and small entities such as ``wall\_plug'', ``cup'', and ``coaster'' are included. 
While the number of classes is larger than in several common datasets a mapping to other class lists is straightforward.

We publish a minimal Replica C++ SDK with the dataset, that demonstrates how to render the Replica reconstructions. The SDK may be used to inspect the dataset and as a starting point for further development. 
For machine learning applications we recommend the use of the AI
Habitat~\cite{habitat19arxiv} simulator which integrates with PyTorch and allows
rendering from Replica directly into PyTorch Tensors for deep learning. The AI
Habitat simulator supports rendering RGB, depth, semantic instance and semantic
class segmentation images at up to 10$k$ frames per second.

\subsection{Data Organization}

Each Replica dataset scene contains the following data: 
\begin{itemize}
    \item \texttt{mesh.ply}: quad mesh encoding the dense surface of the scene. Each vertex has a color value assigned to it for low resolution and non-HDR rendering of the scene (not recommended). 
    \item \texttt{textures/*}: high dynamic range PTex texture files.
    \item \texttt{glass.sur}: file describing reflectors in the scene. It contains a list of reflector parameter objects. Each reflector is described by the transformation from world coordinates to the reflector plane, a polygon in the reflector plane, a surface normal and the reflectance value. A reflectance of $1$ signals a mirror and anything else a partially transparent glass surface. 
    \item \texttt{semantic.json} and \texttt{semantic.bin}: semantic segmentation of the reconstruction. 
    \item \texttt{preseg.json} and \texttt{preseg.bin}: planar/non-planar segmentation of the reconstruction.
    \item \texttt{habitat}: data exported for use with AI Habitat.
    \begin{itemize}
        \item {\texttt{mesh\_semantic.ply}}: quad mesh with semantic instance ids for each primitive. The class of each instance can be looked up in the \texttt{semantic.json} file in the \texttt{habitat} folder.
        \item \texttt{mesh\_semantic.navmesh}: occupancy information needed for AI Habitat agent simulation.
        \item \texttt{semantic.json}: mapping from a semantic instance id stored with every primitive in \texttt{mesh\_semantic.ply} to the semantic class name.
    \end{itemize}
\end{itemize}

The \texttt{semantic.json} and the \texttt{preseg.json} files represent a segmentation forest data structure by specifying a list of nodes with class names, a list of children and a parent field. Each node has a unique id and is addressed via this id. The corresponding \texttt{semantic.bin} and \texttt{preseg.bin} files contain the list of primitive ids corresponding to each node.

\section{Conclusion}

The Replica dataset sets a new standard for texture, geometry and semantic
resolution as well as quality for reconstruction-based 3D datasets. 
It introduces HDR textures and renderable reflector information.
As such it enables AI agent and ML research that needs access to data beyond
static datasets consisting of collections of images such as ImageNet and COCO. 
Furthermore, due to its realism, it can serve as a generative model for
benchmarking 3D perception systems such as SLAM and dense reconstruction
systems as well as to facilitate research into AR and VR telepresence.

\subsection*{Acknowledgments}

The Replica dataset would not have been possible without the hard work and contributions of Matthew Banks,
Christopher Dotson,
Rashad Barber, Justin Blosch, 
Ethan Henderson,
Kelley Greene,
Michael Thot,
Matthew Winterscheid,
Robert Johnston,
Abhijit Kulkarni,
Robert Meeker,
Jamie Palacios,
Tony Phan,
Tim Petrvalsky,
Sayed Farhad Sadat,
Manuel Santana,
Suruj Singh,
Swati Agrawal, and Hannah Woolums.

\bibliographystyle{plain}
\bibliography{main}

\begin{thebibliography}{10}

\bibitem{Anderson2018-Evaluation}
Peter Anderson, Angel~X. Chang, Devendra~Singh Chaplot, Alexey Dosovitskiy,
  Saurabh Gupta, Vladlen Koltun, Jana Kosecka, Jitendra Malik, Roozbeh
  Mottaghi, Manolis Savva, and Amir~Roshan Zamir.
\newblock On evaluation of embodied navigation agents.
\newblock {\em arXiv:1807.06757}, 2018.

\bibitem{Anderson2018-Language}
Peter Anderson, Qi~Wu, Damien Teney, Jake Bruce, Mark Johnson, Niko
  S{\"u}nderhauf, Ian Reid, Stephen Gould, and Anton van~den Hengel.
\newblock Vision-and-language navigation: Interpreting visually-grounded
  navigation instructions in real environments.
\newblock In {\em CVPR}, 2018.

\bibitem{antol_iccv15}
Stanislaw Antol, Aishwarya Agrawal, Jiasen Lu, Margaret Mitchell, Dhruv Batra,
  C.~Lawrence Zitnick, and Devi Parikh.
\newblock {VQA: Visual Question Answering}.
\newblock In {\em ICCV}, 2015.

\bibitem{armenicvpr16}
Iro Armeni, Ozan Sener, Amir~R. Zamir, Helen Jiang, Ioannis Brilakis, Martin
  Fischer, and Silvio Savarese.
\newblock {3D} semantic parsing of large-scale indoor spaces.
\newblock In {\em CVPR}, 2016.

\bibitem{burley2008ptex}
Brent Burley and Dylan Lacewell.
\newblock Ptex: Per-face texture mapping for production rendering.
\newblock In {\em Computer Graphics Forum}, volume~27, pages 1155--1164. Wiley
  Online Library, 2008.

\bibitem{Chang2017}
Angel Chang, Angela Dai, Thomas Funkhouser, Maciej Halber, Matthias Niessner,
  Manolis Savva, Shuran Song, Andy Zeng, and Yinda Zhang.
\newblock {Matterport3D}: Learning from {RGB-D} data in indoor environments.
\newblock In {\em 3DV}, 2017.
\newblock https://niessner.github.io/Matterport/.

\bibitem{craik43}
Kenneth J.~W. Craik.
\newblock {\em The Nature of Explanation}.
\newblock Cambridge University Press, 1943.

\bibitem{dai2017scannet}
Angela Dai, Angel~X. Chang, Manolis Savva, Maciej Halber, Thomas Funkhouser,
  and Matthias Nie{\ss}ner.
\newblock Scannet: Richly-annotated {3D} reconstructions of indoor scenes.
\newblock In {\em CVPR}, 2017.
\newblock http://www.scan-net.org/.

\bibitem{embodiedqa}
Abhishek Das, Samyak Datta, Georgia Gkioxari, Stefan Lee, Devi Parikh, and
  Dhruv Batra.
\newblock {E}mbodied {Q}uestion {A}nswering.
\newblock In {\em CVPR}, 2018.

\bibitem{engel2017direct}
Jakob Engel, Vladlen Koltun, and Daniel Cremers.
\newblock Direct sparse odometry.
\newblock {\em TPAMI}, 40(3):611--625, 2017.

\bibitem{felzenszwalb2004efficient}
Pedro~F Felzenszwalb and Daniel~P Huttenlocher.
\newblock Efficient graph-based image segmentation.
\newblock {\em IJCV}, 59(2):167--181, 2004.

\bibitem{2012-scenesynth}
Matthew Fisher, Daniel Ritchie, Manolis Savva, Thomas Funkhouser, and Pat
  Hanrahan.
\newblock Example-based synthesis of {3D} object arrangements.
\newblock In {\em ACM SIGGRAPH Asia}, 2012.

\bibitem{garcia2018robotrix}
Alberto Garcia-Garcia, Pablo Martinez-Gonzalez, Sergiu Oprea, John~Alejandro
  Castro-Vargas, Sergio Orts-Escolano, Jose Garcia-Rodriguez, and Alvaro
  Jover-Alvarez.
\newblock The robotrix: An extremely photorealistic and very-large-scale indoor
  dataset of sequences with robot trajectories and interactions.
\newblock In {\em IROS}, pages 6790--6797. IEEE, 2018.

\bibitem{handa1511scenenet}
A~Handa, V~Patraucean, V~Badrinarayanan, S~Stent, and R~Cipolla.
\newblock Scenenet: understanding real world indoor scenes with synthetic data.
  arxiv preprint (2015).
\newblock {\em arXiv preprint arXiv:1511.07041}, 2015.

\bibitem{Jakob2015Instant}
Wenzel Jakob, Marco Tarini, Daniele Panozzo, and Olga Sorkine-Hornung.
\newblock Instant field-aligned meshes.
\newblock {\em ACM Transactions on Graphics}, 34(6), November 2015.

\bibitem{krizhevsky_nips12}
Alex Krizhevsky, Ilya Sutskever, and Geoff Hinton.
\newblock {ImageNet} classification with deep convolutional neural networks.
\newblock In {\em NIPS}, 2012.

\bibitem{InteriorNet18}
Wenbin Li, Sajad Saeedi, John McCormac, Ronald Clark, Dimos Tzoumanikas, Qing
  Ye, Yuzhong Huang, Rui Tang, and Stefan Leutenegger.
\newblock Interiornet: Mega-scale multi-sensor photo-realistic indoor scenes
  dataset.
\newblock In {\em BMVC}, 2018.

\bibitem{liepa2003filling}
Peter Liepa.
\newblock Filling holes in meshes.
\newblock In {\em ACM SIGGRAPH Symposium on Geometry Processing}, pages
  200--205, 2003.

\bibitem{mscoco}
Tsung-Yi Lin, Michael Maire, Serge Belongie, James Hays, Pietro Perona, Deva
  Ramanan, Piotr Dollár, and C.~Lawrence Zitnick.
\newblock Microsoft {COCO}: Common objects in context.
\newblock In {\em ECCV}, 2014.

\bibitem{Lorensen1987MC}
William~E. Lorensen and Harvey~E. Cline.
\newblock Marching cubes: A high resolution {3D} surface construction
  algorithm.
\newblock In {\em Proceedings of the 14th Annual Conference on Computer
  Graphics and Interactive Techniques}, SIGGRAPH '87, pages 163--169, New York,
  NY, USA, 1987. ACM.

\bibitem{cgalpmp2019}
S{\'e}bastien Loriot, Jane Tournois, and Ilker~O. Yaz.
\newblock Polygon mesh processing.
\newblock In {\em {CGAL} User and Reference Manual}. {CGAL Editorial Board},
  {4.14} edition, 2019.

\bibitem{mur2015orb}
Raul Mur-Artal, Jose Maria~Martinez Montiel, and Juan~D Tardos.
\newblock {ORB-SLAM}: a versatile and accurate monocular {SLAM} system.
\newblock {\em TRO}, 31(5):1147--1163, 2015.

\bibitem{newcombe2011kinectfusion}
Richard~A Newcombe, Shahram Izadi, Otmar Hilliges, David Molyneaux, David Kim,
  Andrew~J. Davison, Pushmeet Kohi, Jamie Shotton, Steve Hodges, and Andrew
  Fitzgibbon.
\newblock Kinectfusion: Real-time dense surface mapping and tracking.
\newblock In {\em 2011 IEEE International Symposium on Mixed and Augmented
  Reality}, pages 127--136. IEEE, 2011.

\bibitem{habitat19arxiv}
Manolis {Savva*}, Abhishek {Kadian*}, Oleksandr {Maksymets*}, Yili Zhao, Erik
  Wijmans, Bhavana Jain, Julian Straub, Jia Liu, Vladlen Koltun, Jitendra
  Malik, Devi Parikh, and Dhruv Batra.
\newblock Habitat: A platform for embodied ai research.
\newblock {\em arXiv preprint arXiv:1904.01201}, 2019.

\bibitem{Song2017}
Shuran Song, Fisher Yu, Andy Zeng, Angel~X Chang, Manolis Savva, and Thomas
  Funkhouser.
\newblock Semantic scene completion from a single depth image.
\newblock In {\em CVPR}, 2017.

\bibitem{sutton81}
R.~S. Sutton and A.~G. Barto.
\newblock An adaptive network that constructs and uses an internal model of its
  world.
\newblock {\em Cognition and Brain Theory}, 1981.

\bibitem{whelan2018reconstructing}
Thomas Whelan, Michael Goesele, Steven~J. Lovegrove, Julian Straub, Simon
  Green, Richard Szeliski, Steven Butterfield, Shobhit Verma, and Richard
  Newcombe.
\newblock Reconstructing scenes with mirror and glass surfaces.
\newblock {\em ACM Transactions on Graphics (TOG)}, 37(4):102, 2018.

\bibitem{Xia2018}
Fei Xia, Amir~R. Zamir, Zhiyang He, Alexander Sax, Jitendra Malik, and Silvio
  Savarese.
\newblock Gibson env: Real-world perception for embodied agents.
\newblock In {\em CVPR}, 2018.
\newblock http://gibsonenv.stanford.edu/database/.

\end{thebibliography}

\end{document}